\documentclass[runningheads]{llncs}
\usepackage[T1]{fontenc}
\usepackage{hyperref}
\usepackage{times}
\usepackage{epsfig}
\usepackage{graphicx}
\usepackage{amsmath}
\usepackage{amssymb}
\usepackage{booktabs}
\usepackage{multirow}
\usepackage{gensymb}
\usepackage{bbding}
\usepackage{soul}
\usepackage{multirow}
\usepackage{array}
\usepackage{enumitem}
\usepackage{pifont}
\usepackage{graphicx}
\usepackage{subcaption}
\usepackage{colortbl}
\usepackage{placeins}
\usepackage[toc,title,page]{appendix}

\usepackage{graphicx}
\usepackage{xcolor}
\usepackage{amssymb}
\usepackage{wrapfig}
\usepackage{bm}
\usepackage{tikz,pgfplots}
\pgfplotsset{compat=1.17}
\usepackage{afterpage}
\usepackage[ruled,vlined]{algorithm2e}
\usepackage{pifont}
\usepackage{cleveref}   
\newcommand{\xmark}{\ding{55}}%

\newcommand\blfootnote[1]{%
  \begingroup
  \renewcommand\thefootnote{}\footnote{#1}%
  \addtocounter{footnote}{-1}%
  \endgroup
}

\makeatletter
\DeclareRobustCommand\onedot{\futurelet\@let@token\@onedot}

\def\@onedot{\ifx\@let@token.\else.\null\fi\xspace}
\def\eg{\emph{e.g}\onedot} 
 
\def\cf{\emph{c.f}\onedot}

\newcommand{\boldparagraph}[1]{\noindent{\bf #1}}
\newcommand{\figref}[1]{Figure~\ref{#1}}
\newcommand{\secref}[1]{Section~\ref{#1}}

\newcommand{\tabref}[1]{Table~\ref{#1}}
\begin{document}
\title{NVSMask3D: Hard Visual Prompting with Camera Pose Interpolation for 3D
Open Vocabulary Instance Segmentation}
%

\author{Junyuan Fang\inst{1, 2}  \and
Zihan~Wang\inst{1} \and
Yejun~Zhang\inst{1} \and
Shuzhe~Wang\inst{1} \and 
Iaroslav~Melekhov$^{\bullet,}$\inst{1}  \and
Juho~Kannala\inst{1, 3} }
\authorrunning{J. Fang et al.}
%
\institute{Aalto University, Espoo, Finland \and
University of Helsinki, Helsinki, Finland \and
University of Oulu, Oulu, Finland \\
\email{\{junyuan.fang, zihan.1.wang, yejun.zhang, shuzhe.wang, iaroslav.melekhov, juho.kannala\}@aalto.fi}} 

\titlerunning{NVSMask3D}
%
\maketitle              

\blfootnote{$^\bullet$ The work was done prior to joining Amazon}

\begin{abstract}
Vision-language models (VLMs) have demonstrated impressive zero-shot transfer capabilities in image-level visual perception tasks. However, they fall short in 3D instance-level segmentation tasks that require accurate localization and recognition of individual objects. To bridge this gap, we introduce a novel 3D Gaussian Splatting based hard visual prompting approach that leverages camera interpolation to generate diverse viewpoints around target objects without any 2D-3D optimization or fine-tuning. Our method simulates realistic 3D perspectives, effectively augmenting existing hard visual prompts by enforcing geometric consistency across viewpoints. This training-free strategy seamlessly integrates with prior hard visual prompts, enriching object-descriptive features and enabling VLMs to achieve more robust and accurate 3D instance segmentation in diverse 3D scenes. Our code will be released in \href{https://github.com/junyuan-fang/nvsmask3d}{https://github.com/junyuan-fang/nvsmask3d}

\keywords{3D Scene understanding \and Open vocabulary \and 3D Gaussian splatting.}
\end{abstract}

\section{Introduction}
Semantic segmentation of 3D scenes holds immense research value due to its broad applications in robot navigation, object localization, autonomous driving, augmented and virtual reality. 
Traditional 3D segmentation models~\cite{qi2017pointnet,qi2017pointnet++} require large, expensive annotated datasets, limiting scalability and adaptability to unseen categories.
Open-vocabulary 3D segmentation~\cite{peng2023openscene} aims to address this limitation by enabling models to recognize and segment a diverse range of objects, including those with long-tail distributions. However, developing effective open-vocabulary models faces significant challenges due to the scarcity of large-scale, diverse 3D segmentation datasets. 
\begin{figure}[t]
    \centering
    \vspace{-10pt}
    \begin{minipage}{0.40\textwidth}
        \centering
        \includegraphics[width=\linewidth]{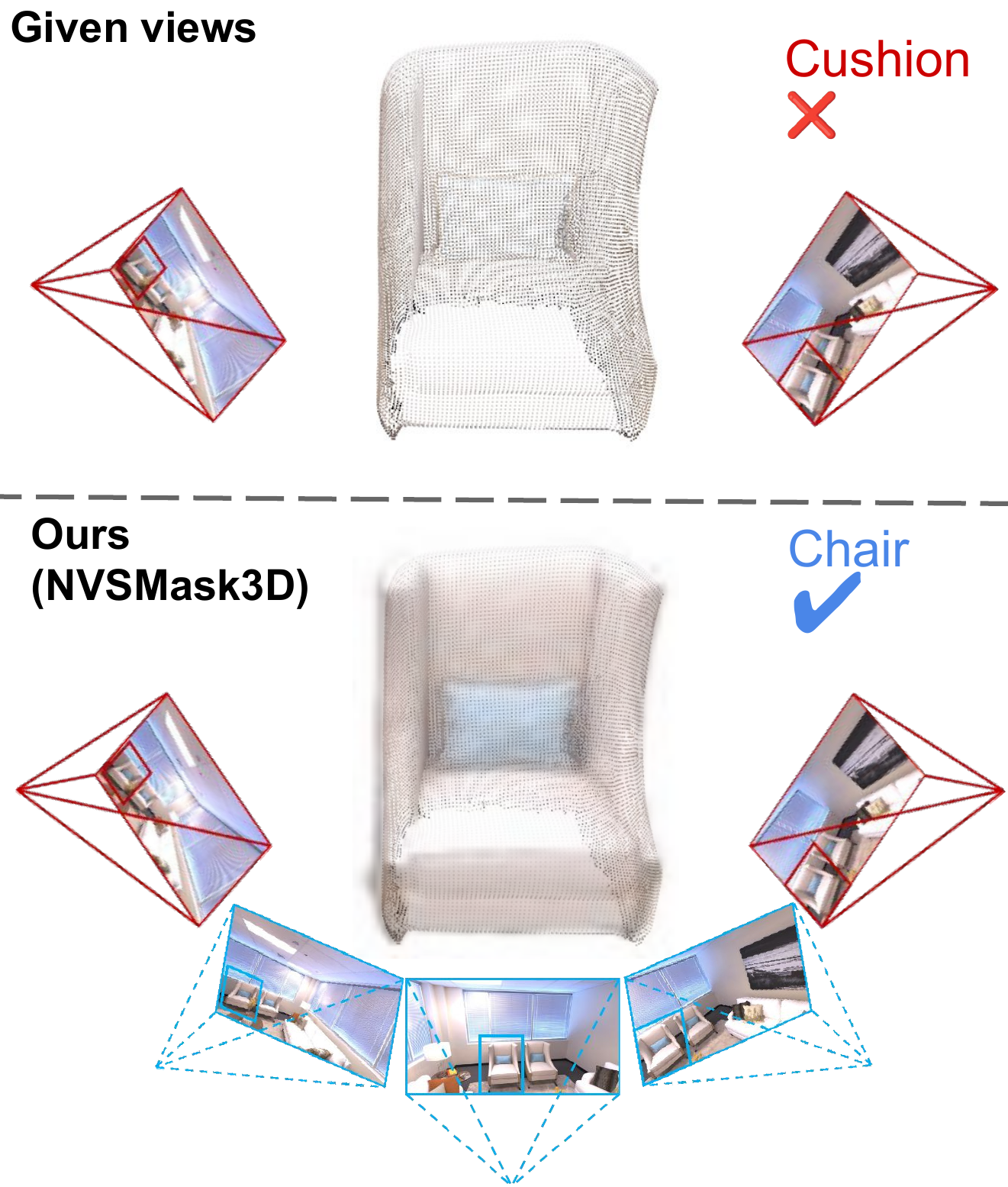}
    \end{minipage}%
    \hfill
    \begin{minipage}{0.55\textwidth}
                \caption{\textbf{Upper:} Previous CLIP-based back-projection techniques~\cite{takmaz2023openmask3d,lu2023ovir,nguyen2023open3dis} struggle with object misclassification in complex 3D scenes. Limited views here cause the chair to be incorrectly identified as a "cushion" when both objects appear together. \textbf{Lower:} NVSMask3D utilizes novel view synthesis with hard visual prompts to generate interpolated views around the target object, creating a more continuous and detailed 3D representation.}
        \label{fig:teaser}
    \end{minipage}
    \vspace{-10pt}
\end{figure}

Current methods for 3D open-vocabulary scene understanding can be categorized into three main types. The first category, 2D knowledge distillation into a 3D backbone, includes works like~\cite{peng2023openscene,zhang2023clip,chen2023clip2scene}, which freeze the visual encoder of 2D vision-language models (VLMs) to transfer 2D visual-language features from pre-trained 2D models into a 3D neural network. The second category~\cite{ding2023pla,yang2024regionplc,ding2024lowis3d} leverages 2D captioning models, where the text encoder of a VLM is frozen to guide the training of a 3D backbone using captions generated from 2D images. The third category focuses on mapping 2D features into 3D space by utilizing known camera poses. CLIP-based back-projection techniques~\cite{nguyen2024open3dis,takmaz2023openmask3d,lu2023ovir} assign 2D CLIP features to 3D points through back-projection, enabling 3D point-level object retrieval. Similarly, 2D open-world detectors, such as YOLOWorld \cite{Cheng2024YOLOWorld}, identify and segment regions with high accuracy directly in 2D then projecting these segmented regions into 3D space, as demonstrated in works like OpenYOLO3D and OpenIns3D \cite{boudjoghra2024openyolo,huang2024openins3d}. 

These introduced zero-shot approaches primarily rely on improvements in 2D~\cite{takmaz2023openmask3d,peng2023openscene} and 3D~\cite{nguyen2024open3dis,huang2024openins3d} mask quality, as well as advancements in VLMs~\cite{boudjoghra2024openyolo,huang2024openins3d}, to enhance 3D open-vocabulary understanding. However, their reliance on fixed multi-scale, multi-view 2D captures from meshes, point clouds, or RGB-D images limits either the diversity of mesh data~\cite{huang2024openins3d} or the flexibility of fixed camera poses~\cite{nguyen2024open3dis,takmaz2023openmask3d,lu2023ovir} provided by the dataset. As a result, these sparse and object-side views fail to capture a complete and detailed representation of the object (\cf~\figref{fig:teaser}).

Inspired by the effectiveness of text prompt engineering~\cite{radford2021learning,pham2023combined,zhai2022lit} in enriching semantic information and boosting VLM performance through more descriptive prompts, we propose a similar approach by incorporating additional visual prompts to address the Open Vocabulary Instance Segmentation task. To achieve this, we leverage Novel View Synthesis (NVS) with 3D Gaussian Splatting (GS) to generate interpolated views around target objects, treating these as an additional visual prompt. Our experiments on the Replica and ScanNet++ datasets validate this approach, showing that our external visual prompting strategy enriches model perception, enabling VLM to recognize and interpret objects more effectively from diverse viewpoints. Our main contributions are as follows:

\begin{itemize} 
\item We introduce camera pose interpolation as a visual prompt, generating continuous 3D perspectives to enhance 3D instance segmentation without requiring joint 2D-3D feature optimization or fine-tuning.
\item Our visual prompt integrates seamlessly with other visual prompting techniques. When combined with our Weighted Feature Balancing (WFB) mechanism, it effectively reduces noise from interpolated views and ensures high-quality, object-descriptive features across diverse viewpoints.
\item Experiments on multiple datasets demonstrate that NVSMask3D outperforms baseline methods, validating its robustness and effectiveness across varied 3D scenes. \end{itemize}

\section{Related Work}

Pioneering work like OpenScene \cite{peng2023openscene} bridges the 2D and 3D domains by distilling CLIP-based 2D semantic features~\cite{ghiasi2022scaling,li2022language} into 3D models, aligning 2D visual-text embeddings with 3D point cloud segmentation. Similarly, OpenMask3D \cite{takmaz2023openmask3d} utilizes class-agnostic 3D proposals \cite{schult2023mask3d} to separate 3D segmentation and semantic labelling, extracting 2D object crops with SAM \cite{Kirillov_2023_ICCV} and using multiview averaging of CLIP features for object segmentation across views. Other CLIP-based back-projection methods, such as OVIR-3D \cite{lu2023ovir} and Open3DIS \cite{nguyen2024open3dis}, focus on fusing 2D masks \cite{Kirillov_2023_ICCV} into 3D space, with Open3DIS introducing a novel mask fusion strategy that improves accuracy for smaller objects. However, these approaches often depend on better 3D masks \cite{schult2023mask3d,landrieu2019point,robert2023efficient}, 2D masks \cite{Kirillov_2023_ICCV}, or improved 2D instance-level foundational models \cite{Cheng2024YOLOWorld}. In contrast, our method achieves robust performance without relying on external foundational models.

Advancements in 3D scene understanding have also been driven by improved scene representations like Neural Radiance Fields (NeRF) \cite{mildenhall2021nerf} and 3D Gaussian Splatting (3D-GS) \cite{kerbl3Dgaussians}, alongside VLMs like CLIP \cite{radford2021learning}, which enable semantic understanding. Methods such as OpenNeRF \cite{engelmann2024opennerf} and LangSplat \cite{qin2023langsplat} rely on explicit 2D supervision, where pixel- or image-level features (\eg, OpenSeg~\cite{ghiasi2022scaling}, CLIP~\cite{radford2021learning}) guide the joint optimization of 3D representations with 2D-based CLIP features, enforcing alignment between the 2D and 3D domains. Our approach, however, bypasses joint feature optimization. Instead, we enhance 2D VLMs for 3D instance segmentation through only hard visual prompting techniques, leveraging camera interpolation to vanilla 3D-GS to generate viewpoint-conditioned object descriptive visual prompts. This offers a novel opportunity for a hard visual prompt-based approach in 3D instance segmentation, serving as an alternative to direct 2D supervision in 3D optimization strategies and demonstrating competitive performance against our baseline.

Beyond direct 2D supervision, visual prompt engineering has emerged as an effective technique for adapting VLMs to downstream tasks. Visual prompting can be categorized into soft and hard prompts. Soft prompts, such as visual prompt tuning \cite{jia2022visual,yoo2023improving}, involve learnable embeddings that are directly incorporated into the model, typically without altering the visual input itself. These learnable prompts are optimized during training and act as internal guides to influence the model's attention. They allow the integration of learnable visual prompts into models, enabling them to adapt flexibly to various downstream tasks  \cite{liang2023open,minderer2024scaling}, improving performance without extensive retraining. In contrast, hard prompts explicitly modify the visual input to direct model attention, using techniques like circles, boxes, Gaussian blur, masks, and cropping \cite{shtedritski2023does,wang2023caption,yang2023fine,kim2023dense}. Unlike soft prompts, hard visual prompts are inherently training-free, making them efficient tools for inference-time adaptation. However, these methods have been largely confined to 2D tasks. In this work, we extend hard visual prompting to the 3D domain by leveraging external viewpoint-conditioned prompts via camera interpolation. Our approach bridges the gap between traditional 2D prompting techniques and 3D open-vocabulary instance segmentation, offering a novel adaptation of hard prompts to improve 3D open vocabulary scene understanding without requiring model fine-tuning or 2D open vocabulary feature supervision in 3D optimization.

\section{Method}
\begin{figure*}[t!]
    \centering
    \includegraphics[width=\textwidth]{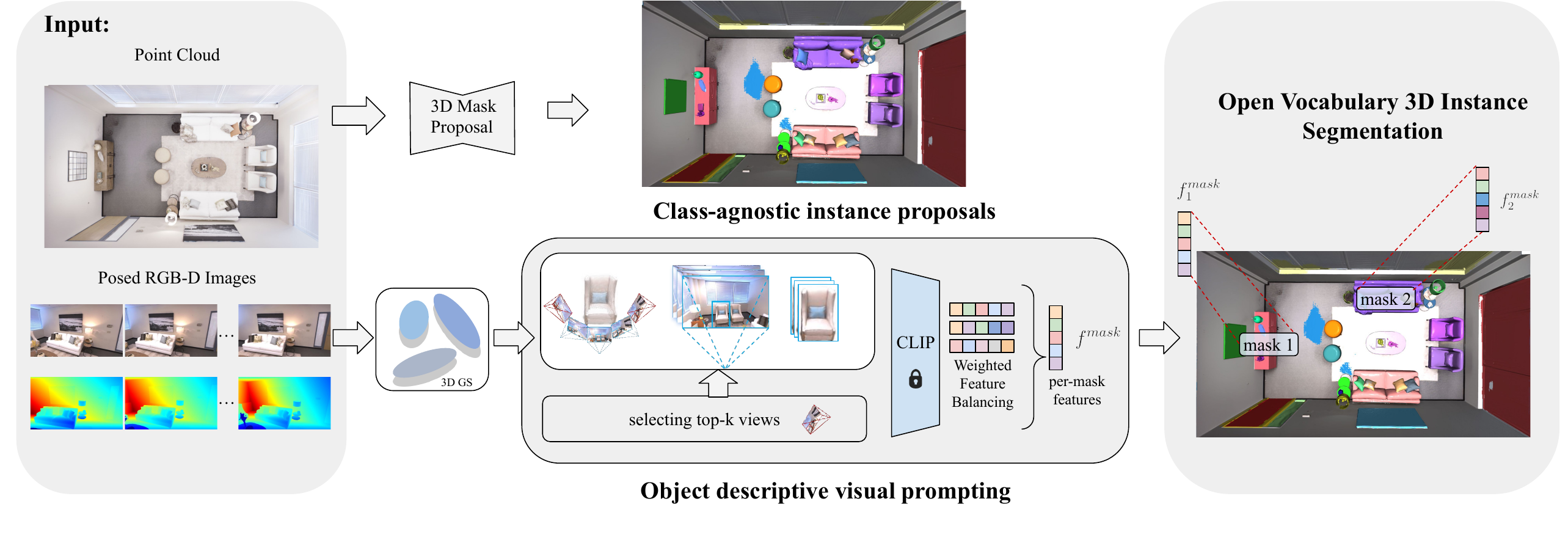}
    \caption{\textbf{Overview of the NVSMask3D Pipeline.} NVSMask3D begins by applying a class-agnostic 3D proposal to segment the input point cloud, generating initial instance masks. Next, the 3D scene is represented using 3D-GS. Top-$k$ camera poses are selected as references to interpolate additional views. Visual prompts are applied to these interpolated views to emphasize object-descriptive features. CLIP features are then extracted from each image and combined using a WFB mechanism, which stabilizes feature contributions from NVS-generated and top-$k$ views, ultimately enhancing 3D instance segmentation accuracy.}
\label{fig:NVSMask3D_pipeline}
\vspace{-10pt}
\end{figure*}


Given a 3D point cloud, posed RGB-D images, and known camera calibrations, our objective is to perform open-vocabulary 3D instance segmentation in a zero-shot manner by leveraging hard visual prompts within our pipeline. As illustrated in~\figref{fig:NVSMask3D_pipeline}, we begin by applying a class-agnostic 3D mask proposal \cite{schult2023mask3d} to segment the point cloud into distinct object instances. Simultaneously, we represent the 3D scene using 3D-GS \cite{kerbl3Dgaussians}, initializing Gaussians at each 3D point with fixed means that align with the object instances, allowing the same instance masks to be used for segmenting Gaussians. For each segmented object, we select descriptive camera poses and determine the object's centroid, allowing us to interpolate object descriptive camera views that ensure the object remains centered across perspectives, which can effectively enhance feature diversity. However, the quality of NVS-generated images is generally lower than that of the original images in the dataset, introducing additional noise.
To enhance feature diversity while mitigating the impact of feature noise, 
we employ weighted feature balancing (WFB) to fuse features from both selected reference views and interpolated views. Notably, we do not optimize the 3D mask itself. Instead, our focus is on generating more continuous and object-descriptive visual prompts, which are then assigned to the VLM. This combination of NVS and our feature fusion prior WFB enhances both segmentation accuracy and object recognition across diverse 3D scenes.

\subsection{Class-agnostic instance proposal} 
Our method begins by generating \( M \) class-agnostic 3D mask proposals, denoted as \( \{ m_j^{3D} \}_{j=1}^{M} \). Each mask \( m_j^{3D} \in \{0, 1\}^N \) is a binary vector over the \( N \) points of the input point cloud \( \mathcal{P} \in \mathbb{R}^{3 \times N} \), where \( m_{j,n}^{3D} = 1 \) indicates that the \( n \)-th point belongs to the \( j \)-th object instance. We generate these masks using the mask-generation module from a pre-trained 3D instance segmentation model, such as Mask3D~\cite{schult2023mask3d}. Unlike Mask3D, which assigns predefined class labels, we instead associate each mask with 3D object-descriptive CLIP features, enabling open-vocabulary instance-level analysis.

\subsection{3D scene representation }

Following vanilla 3D Gaussian Splatting \cite{kerbl3Dgaussians}, we define a Gaussian as 
\(\mathcal{G}_i \{\boldsymbol{\mu_i}, \boldsymbol{\Sigma_i}, o_i, \boldsymbol{c_i} \}\), 
where \(\boldsymbol{\mu_i} \in \mathbb{R}^3\) represents the Gaussian mean, 
\(\boldsymbol{\Sigma}_i \in \mathbb{R}^{3\times3}\) is the covariance matrix, 
\(o_i \in \mathbb{R}\) denotes the opacity, and 
\(\boldsymbol{c}_i \in \mathbb{R}^3\) corresponds to the view-dependent color. 
To render an individual view from the 3D Gaussian scene representation, each Gaussian is projected onto the 2D image plane using the candidate camera pose \(\mathbf{P}\). 
A tile-based rasterizer is employed to improve rendering efficiency by processing Gaussians within localized screen-space tiles. 
The Gaussians in each tile are sorted by z-depth and alpha-blended to compute the final pixel color \(\hat{C}(p)\):
\begin{equation}
\hat{C}(p) = \sum_{i\in \mathcal{N}} c_i \alpha_i T_i, \quad \text{where} \quad  T_i = \prod_{j=1}^{i-1} (1 - \alpha_j),
\label{eq:color_blending}
\end{equation}
where \(\mathcal{N}\) denotes the set of Gaussians within a given tile, and \(T_i\) represents the accumulated transmittance at the rendered pixel \(p\), ensuring correct depth-aware compositing. 
The blending weight \(\alpha_i\) for the \(i\)th Gaussian is computed in view space based on its projected mean \(\boldsymbol{\hat{\mu}_i}\):
\begin{equation}
\alpha_i = o_i \cdot \exp\left( -\frac{1}{2} (p - \boldsymbol{\hat{\mu}_i})^\top \Sigma_i^{-1} (p - \boldsymbol{\hat{\mu}_i}) \right).
\label{eq:alpha_weight}
\end{equation}
During optimization, we initialize and anchor \(\boldsymbol{\mu_i}\) using the given point cloud \(\mathcal{P}\). 
We then optimize all parameters of \(\mathcal{G}_i\) except for the means \(\boldsymbol{\mu_i}\) and disable the Adaptive Density Control stage. 
This ensures a high-fidelity scene representation for NVS while preserving the original correspondence between the point cloud and the Gaussian means. 

After optimization, novel views are synthesized by rendering Gaussians \(\mathcal{G}_i\) using interpolated camera poses \( \mathbf{P} \). These interpolated viewpoints, obtained via the Object descriptive visual prompting method described in Section~\ref{subsec:visual_prompt}, facilitate smooth and continuous exploration beyond the original dataset poses. The rendering follows Equation~\eqref{eq:color_blending}, where interpolated Gaussians are projected, rasterized, and alpha-blended into the final image.


\subsection{Object descriptive visual prompting} \label{subsec:visual_prompt}
In order to assign object-descriptive CLIP features to \( m_j^{3D} \), we propose a zero-shot hard visual prompting method based on object-descriptive camera interpolation. First, we compute the camera pose visibility score to identify the most relevant reference poses for \(m_j^{3D}\). Next, we adjust the camera's viewing direction to ensure the target object is centered in the image. Then, we apply linear interpolation and camera calibration to generate intermediate camera poses, ensuring the object remains visible and centered. This step minimizes the risk of losing important visual details and enhances prompt effectiveness, especially in zero-shot scenarios where consistent views improve model robustness and accuracy.


\begin{figure*}[t!]
    \centering
    \begin{subfigure}{0.3\textwidth}
        \centering
        \includegraphics[width=\textwidth, trim=185 50pt 200 120pt, clip]{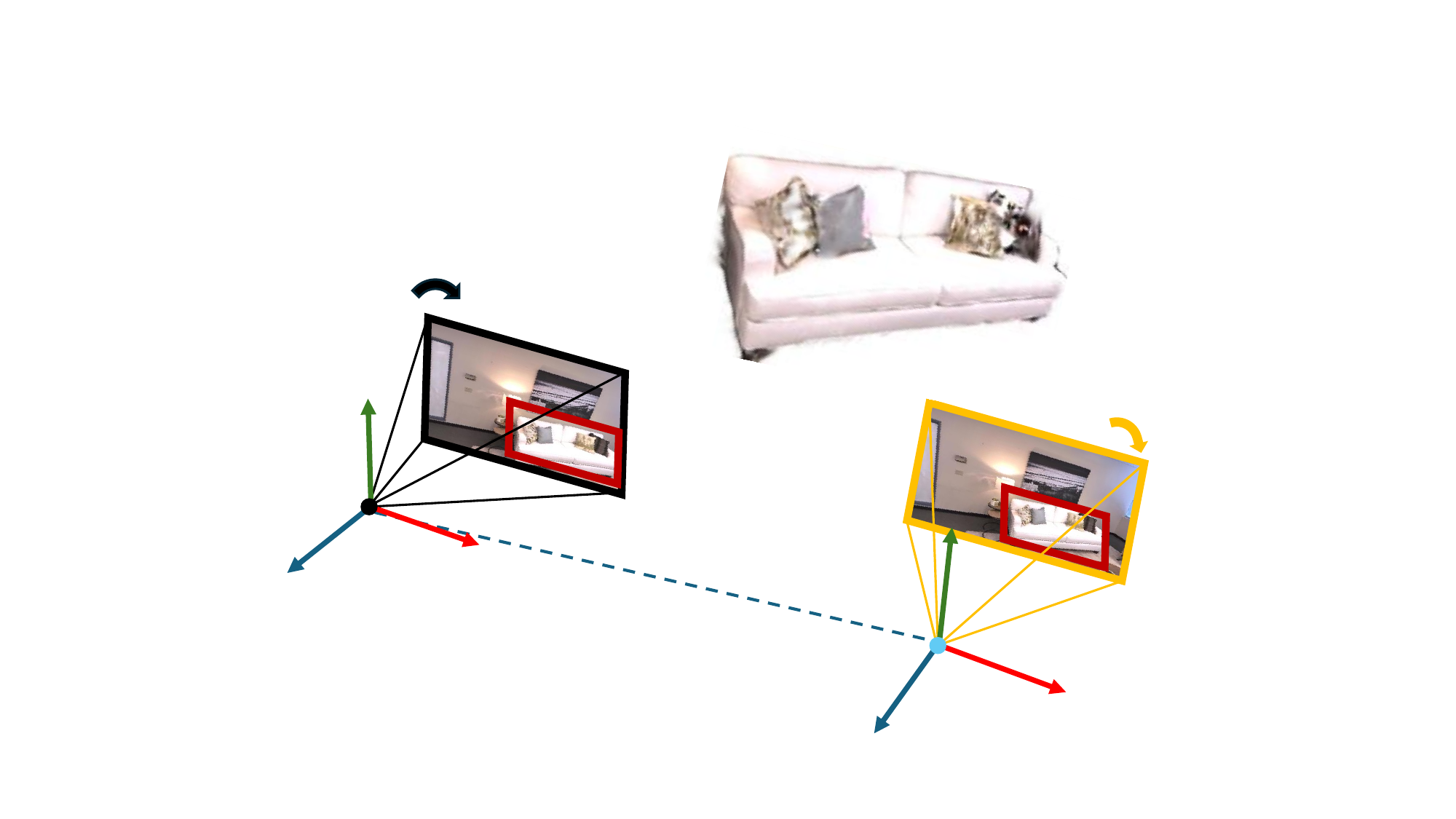} 
    \end{subfigure}
    \hfill
    \begin{subfigure}{0.3\textwidth}
        \centering
        \includegraphics[width=\textwidth, trim=185 50pt 200 120pt, clip]{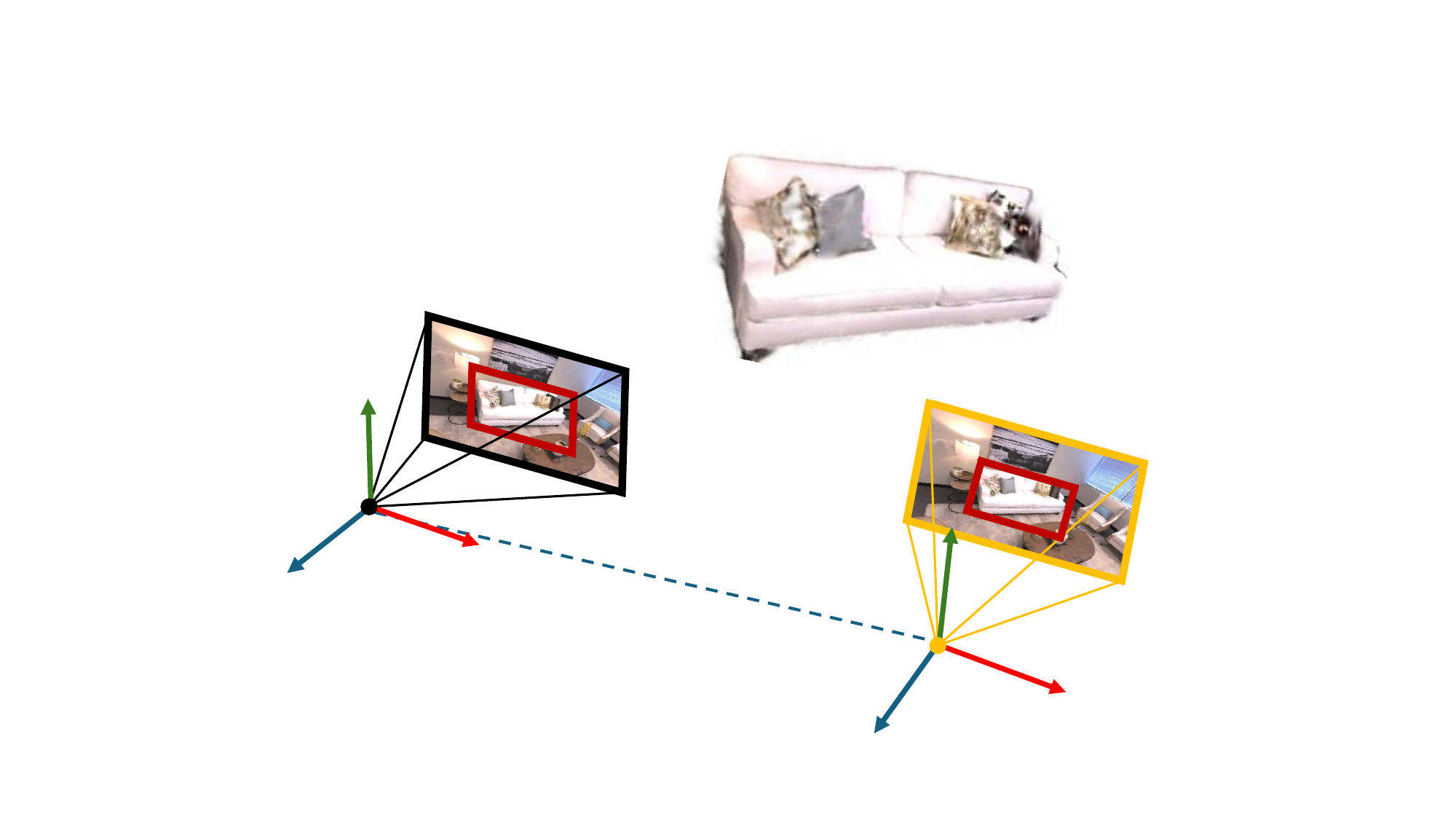} 
    \end{subfigure}
    \hfill
    \begin{subfigure}{0.3\textwidth}
        \centering
        \includegraphics[width=\textwidth, trim=185 50pt 200 120pt, clip]{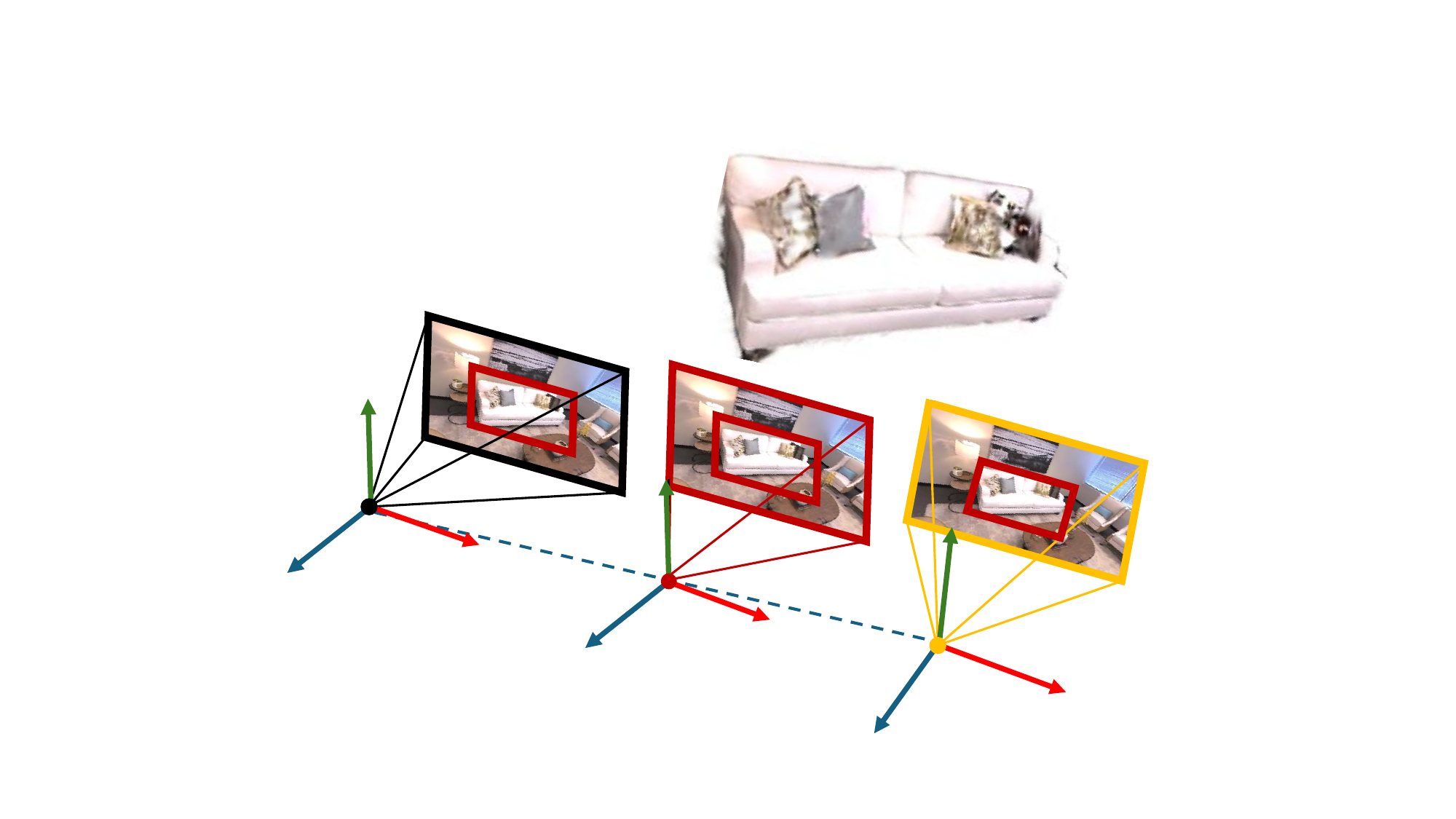} 
    \end{subfigure}
    
    \caption{Reference pose, camera pose adjustment and camera pose interpolation. 
    Initial camera poses before adjustment, where the camera pose is misaligned with the object center (left). Adjusted camera pose, realigned towards the object's geometric center (middle). Interpolation and final readjustment of the interpolated camera pose to generate novel views from 3D-GS (right).}  
    \label{fig:visual_prompt}
    \vspace{-10pt}
\end{figure*}

\subsubsection{Camera pose visibility score.} \label{subsec:camera_visibility_score}
To select the most descriptive camera poses, we compute the Camera Pose Visibility Score \( S_{ij} \) \cite{takmaz2023openmask3d} for each candidate camera pose \( \mathbf{P}_i \) and 3D object \( \mathbf{p}_j = m_j^{3D} \odot \mathcal{P} \). This score quantifies how effectively a camera pose captures the object's features by considering the number of object points visible from the pose that are not occluded. For a set of 3D points \( \{ \mathbf{p}_j \}_{j=1}^N \), where \( \mathbf{p}_j = \left[ p_{j,x}, p_{j,y}, p_{j,z} \right] \) represents each point in world coordinates, we project these points onto the image plane using the intrinsic matrix \( K \) and the world-to-camera transformation \( \left[ \mathbf{R}_i \mid \mathbf{t}_i \right] \), where \( \mathbf{R}_i \) is the rotation matrix, defining the orientation of the camera in world coordinates, and \( \mathbf{t}_i \) is the translation vector, specifying the position of the camera in world coordinates:

\begin{equation}
    \begin{bmatrix}
        u_{ij} \\
        v_{ij} \\
        z_{ij}
    \end{bmatrix}
    =
    K \cdot \begin{bmatrix} R_i \mid \mathbf{t}_i \end{bmatrix} \cdot 
    \begin{bmatrix}
        \mathbf{p}_j \\
        1
    \end{bmatrix}
    =
    K \cdot 
    \begin{bmatrix}
        p_{j,x} \\
        p_{j,y} \\
        p_{j,z} \\
        1
    \end{bmatrix},
\end{equation}
where \( u_{ij}, v_{ij}, z_{ij} \) 
are the pixel coordinates and depth value. A point \( \mathbf{p}_j \) is considered visible from pose \( i \) if it satisfies:
\[
0 \leq u_{ij} < W, \quad 0 \leq v_{ij} < H, \quad z_{ij} > 0, \quad | z_{ij} - d_{ij} | \leq \delta.
\]
Here, \( W \) and \( H \) are the image width and height, \( d_{ij} \) is the depth map value at \( (u_{ij}, v_{ij}) \), and \( \delta \) is a predefined depth threshold. The visibility indicator \( V_{ij} \) is set to 1 if the point \( \mathbf{p}_j \) is visible from pose \( \mathbf{P}_i \), and 0 otherwise. The Camera Pose Visibility Score \( S_{ij} \) for each camera pose is then calculated as the total number of visible points \( S_{ij} = \sum_{j} V_{ij}.\) By computing \( S_{ij} \) for all candidate poses, we 
select top-\( k \) camera poses with the highest scores. This ensures that the chosen camera poses captures the most significant aspects of the object with maximal coverage and minimal occlusions.

\boldparagraph{Object-centered camera adjustment.} For each selected camera pose, we compute a unit vector \( \mathbf{v}_c \in \mathbb{R}^{3} \) pointing to the object's center in world coordinates. The center is estimated using the geometric median of the 3D points \( \{ \mathbf{p}_i \}_{i=1}^N \), providing robustness of the central point \( \mathbf{x} \in \mathbb{R}^3 \) against outliers and non-uniform distributions. The geometric median \( \mathbf{c} \in \mathbb{R}^3 \) is defined as the centroid point that minimizes the sum of Euclidean distances to all points in the set:
\begin{equation}
    \mathbf{c} = \arg\min_{\mathbf{x} } \sum_{i=1}^{N} \| \mathbf{p}_i - \mathbf{x} \|_2.
    \label{eq:geometric_median}
\end{equation}
This geometric median \( \mathbf{c} \) in vector form \( \mathbf{v}_c \) is then utilized to adjust the rotation of each camera pose, ensuring that the camera is oriented towards the object center. To adjust the orientation of each camera towards the object center, we employ a transformation akin to the "LookAt" transformation. The objective is to align the camera's principal axis with the computed unit vector \( \mathbf{v}_c \), which directs towards the object geometric median $\mathbf{c}$. This transformation is defined by a rotation matrix \( \mathbf{R}_i \in \mathbb{R}^{3 \times 3} \), which reorients each camera pose to face the target point while preserving a consistent up direction. The original up direction, \( \mathbf{u}_i \), is extracted and normalized from the given camera pose matrix and serves as a reference for constructing the orthogonal basis required to define \( \mathbf{R}_i \). The three orthogonal vectors are computed as follows:

\begin{equation}
\mathbf{f} = \frac{\mathbf{v}_c - \mathbf{t}_i}{\| \mathbf{v}_c - \mathbf{t}_i \|}, \quad  
\mathbf{u}_i = \frac{\mathbf{u}_i}{\| \mathbf{u}_i \|}, \quad
\mathbf{r} = \frac{\mathbf{u}_i \times \mathbf{f}}{\| \mathbf{u}_i \times \mathbf{f} \|}, \quad
\mathbf{u} = \mathbf{f} \times \mathbf{r},
\end{equation}
where $\mathbf{t}_i$ represents the translation vector from the camera pose $\mathbf{P}_i$, $\mathbf{f}$ is the “Forward Vector” (depicted by the blue arrow in~\figref{fig:visual_prompt}), and $\mathbf{u_i}$ is the “Original Up Vector” (shown as the green arrow), The $\mathbf{r}$ is the “Right Vector” (illustrated by the red arrow) is computed as the normalized cross product of $\mathbf{u_i}$ and $\mathbf{f}$, while the “Adjusted Up Vector” $\mathbf{u}$ is obtained by taking the cross product of $\mathbf{f}$ and $\mathbf{r}$ to ensure orthogonality. These orthogonal vectors define the new camera rotation matrix as: \( \mathbf{R}_i = [\mathbf{r} \ \mathbf{u} \ -\mathbf{f}] \). This transformation reorients the camera to face the object's geometric median \( \mathbf{v}_c \), ensuring that the target remains centered in the field of view (FOV).

\boldparagraph{Camera interpolation.} To smoothly interpolate transitions between camera poses to generate consecutive camera views. We begin by linearly interpolating the translation \( \mathbf{t}_{n} \) between two consecutive camera poses \( \mathbf{P}_a \) and \( \mathbf{P}_b \):
\begin{equation}
\mathbf{t}_{n} = (1 - t_n) \mathbf{t}_a + t_n \mathbf{t}_b, \quad t_n = \frac{n}{N_{\text{interp}}
 + 1}, \quad n = 1, \ldots, N_{\text{interp}} ,
\end{equation}
where \( \mathbf{t}_a \) and \( \mathbf{t}_b \) are reference translation vectors from camera poses \( \mathbf{P}_a \) and \( \mathbf{P}_b \), \text{$ t_n$}  represents the interpolation factor between the two translation vectors, and \( N_{\text{interp}} \) is the total number of intermediate steps between \( \mathbf{t}_a \) and \( \mathbf{t}_b \). After determining \(\mathbf{t}_{n}\), the next step is to design the camera rotation matrix \( \mathbf{R}_{n} \) such that the interpolated camera pose \( \mathbf{P}_{n} \) continues to point towards the object center \( \mathbf{v}_c \). To ensure camera Roll axis consistency, the rotation \( \mathbf{R}_{n} \) is computed using the up vector \( \mathbf{u}_a \) from the starting pose \( \mathbf{P}_a \) as an initialization. This guarantees that the interpolated camera maintains a stable up direction throughout the interpolation process, avoiding unwanted flips or abrupt changes in orientations. 
Once both the interpolated translation and rotation are determined, complete camera pose \( \mathbf{P}_{n} \in \mathbb{R}^{4 \times 4} \) is formulated in homogeneous coordinates:
\begin{equation}
\mathbf{P}_n = \begin{bmatrix} 
\mathbf{R}_n & \mathbf{t}_n \\ 
\mathbf{0}^\top & 1 
\end{bmatrix}\label{eq:interpolated_camera}
\end{equation}

\begin{figure*}[t!]
    \centering
    
    \begin{subfigure}{0.20\textwidth}
        \centering
        \captionsetup{ labelformat=empty, font=scriptsize}
        \includegraphics[width=\textwidth, trim=25pt 0pt 25pt 0pt, clip]{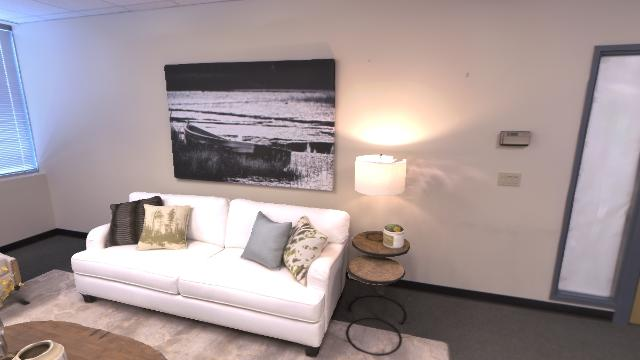} 
        \caption{Raw Image}
    \end{subfigure}
    \hfill
    \begin{subfigure}{0.20\textwidth}
        \centering
        \captionsetup{ labelformat=empty, font=scriptsize}
        \includegraphics[width=\textwidth, trim=25pt 0pt 25pt 0pt, clip]{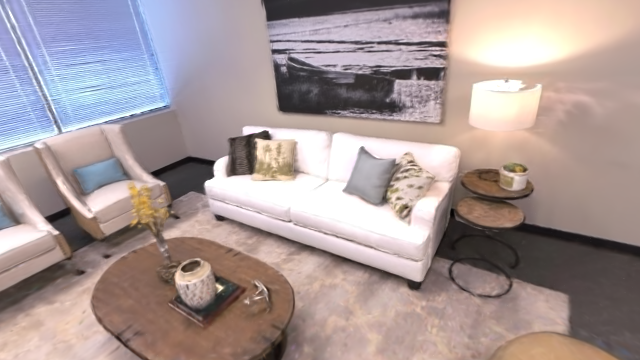} 
        \caption{Camera Interpolation}
    \end{subfigure}
    \hfill
    \begin{subfigure}{0.19\textwidth}
        \centering
        \captionsetup{ labelformat=empty, font=scriptsize}
        \includegraphics[width=\textwidth, trim=0pt 51pt 0pt 51pt, clip]{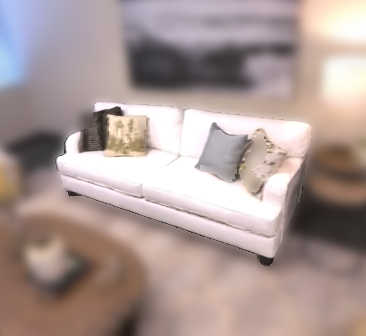} 
        \caption{Blurring}
    \end{subfigure}
    \hfill
    \begin{subfigure}{0.19\textwidth}
        \centering
        \captionsetup{ labelformat=empty, font=scriptsize}
        \includegraphics[width=\textwidth]{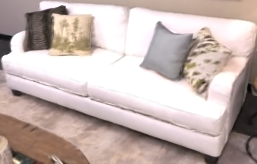} 
        \caption{Cropping}
    \end{subfigure}
    \hfill
    \begin{subfigure}{0.19\textwidth}
        \centering
        \captionsetup{ labelformat=empty, font=scriptsize}

        \includegraphics[width=\textwidth]{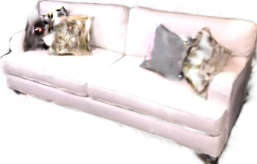} 
        \caption{Segmented Gaussian}
    \end{subfigure}
    \caption{Qualitative renderings of raw images and four different hard visual prompts. From left to right: the raw input image, the interpolated object-centered camera pose, and the subsequent visual prompts: blurring, cropping, and segmented Gaussians.}

    \label{fig:qualitative_results}
    \vspace{-10pt}

\end{figure*}

\boldparagraph{Hard visual prompts.} To improve object descriptive CLIP feature for unlabeled $m_{j}^{3D}$, four different visual prompting techniques can be applied to the view of selected camera poses $\mathbf{P}_i$, as shown in~\figref{fig:qualitative_results}. The first column presents the raw input image, while the remaining four columns illustrate different visual prompting methods: camera interpolation, which ensures the $m_{j}^{3D}$ object remains centered in the interpolated FOV to provide object-descriptive perspectives; blurring \cite{yang2023fine}, which reduces background distraction to focus on the object; cropping \cite{takmaz2023openmask3d}, which isolates the object by removing surrounding regions; and segmented Gaussians, where only the $m_{j}^{3D}$ object's 3D Gaussians are rendered to further exclude background attention. In addition to applying the last three visual prompts to the dataset's raw images, our interpolated camera poses $\mathbf{P}_n$ serve as an external visual prompting technique, providing more descriptive views of each object and enabling integration with the other three 2D-based or 3D-based visual prompts. In this paper, we refer to image modifications such as blurring, cropping, and segmented Gaussians, when applied on top of camera interpolation, as \textit{secondary visual prompts}. 

\boldparagraph{Weighted Feature Balancing (WFB).} Let \( \mathbf{f}_{i}^{\text{CLIP}} \in \mathbb{R}^{D^{\text{CLIP}}} \) denote 2D CLIP features obtained from the top-\( k \) camera poses, and \( \hat{\mathbf{f}}_{j}^{\text{CLIP}} \in \mathbb{R}^{D^{\text{CLIP}}} \) denote the interpolated view's 2D CLIP features from NVS, which can effectively enhance feature diversity. However, the quality of NVS-generated images is generally lower than that of the original camera poses in the dataset, introducing additional noise. The total number of camera poses generated by equation \ref{eq:interpolated_camera} is given by \( \lambda = N_{\text{interp}}(k-1) \).
However, when \( \lambda \) increases beyond \( k \), interpolated \( \hat{\mathbf{f}}_{j}^{\text{CLIP}} \) features start to dominate the representation, potentially introducing excessive noise and diminishing the contribution of original high-quality features, which may degrade performance. Therefore, it is essential to control the influence of interpolations, keeping them below that of the top-\( k \) poses. This limitation makes unweighted feature averaging \cite{peng2023openscene,takmaz2023openmask3d,boudjoghra2024openyolo,nguyen2024open3dis} inappropriate for aggregating point-wise CLIP features \( \mathbf{F}^{\text{CLIP}} \in \mathbb{R}^{D^{\text{CLIP}}} \). Instead, we propose the following weighted feature balancing (WFB) approach:
\begin{equation}
    \mathbf{F}^{\text{CLIP}} = \text{Normalize}\left( \sum_{i=1}^{k} \mathbf{f}_{i}^{\text{CLIP}} + \frac{\alpha}{N_{\text{interp}}
} \sum_{j=1}^{\lambda} \hat{\mathbf{f}}_{j}^{\text{CLIP}} \right).
\label{eq:weighted-feature}
\end{equation}
In this method, we scales the sum of the interpolated features \( \mathbf{f}_{i}^{\text{CLIP}} \) by \( \frac{\alpha}{N_{\text{interp}}} \), where \( \alpha \in (0, 1] \) is the hyperparameter for the feature \( \mathbf{f}_{i}^{\text{CLIP}} \) contribution, add them to the top-\( k \) features, and then normalizes the resulting vector to maintain feature consistency. Finally, we normalize the resulting vector to maintain consistency in the feature space. 
\section{Experiments}
\boldparagraph{Datasets.} We conducted our experiments on a dataset that provides RGB-D images with camera poses, supporting both point cloud instance segmentation and NVS, to validate the effectiveness of our approach. Previous methods~\cite{takmaz2023openmask3d,nguyen2023open3dis,boudjoghra2024openyolo,huang2024openins3d} have primarily focused on the ScanNet200 \cite{rozenberszki2022language} and Replica \cite{straub2019replica} datasets. However, the high motion blur and limited FOV in ScanNet200 make it challenging to optimize a 3D representation capable of generating useful images \cite{yeshwanth2023scannet++}. Therefore, our experiments are primarily conducted on the Replica~\cite{straub2019replica} and ScanNet++ \cite{yeshwanth2023scannet++} datasets. The Replica dataset includes 18 high-quality 3D indoor scene reconstructions, each with dense meshes, high-resolution textures, and semantic annotations, making it ideal for tasks such as NVS and instance segmentation. ScanNet++ provides a large-scale collection of 460 indoor scenes captured with high-end laser scanners and DSLR cameras, offering high-fidelity geometry and color data. For evaluation, we use 8 scenes from the Replica dataset and 50 scenes from the Scannet++ validation set. In ScanNet++, we exclusively utilize COLMAP-aligned poses with their corresponding depth and RGB data.

\boldparagraph{Baselines.} \label{subsubsec:baseline}We compare the proposed approach to two baseline methods based on the state-of-the-art open vocabulary 3D instance segmentation method, OpenMask3D~\cite{takmaz2023openmask3d}: one using 2D SAM~\cite{Kirillov_2023_ICCV} masks and the other relying solely on 3D masks. To ensure fair comparisons, we apply camera interpolation as a visual prompt across both baselines. For both baselines, interpolated views are restricted to use only the projected 3D masks for operations such as cropping, blurring, and segmented Gaussian. This consistent use of 3D masks across interpolated views enables a direct assessment of the additional benefits offered by our object-centric visual prompting techniques in enhancing the CLIP feature representation with WFB for 3D objects. 

\boldparagraph{Evaluation matrics.}We evaluate performance using the standard Average Precision (AP) metric~\cite{dai2017scannet} at fixed IoU thresholds of 25\% and 50\%. Additionally, we compute the mean Average Precision (mAP) across IoU thresholds ranging from 50\% to 95\%, increasing in 5\% intervals. The AP, AP50, and AP25 metrics reported in~\tabref{results} represent averages across different scenes at the scene level.

\begin{table*}[t!] 
\centering 
\resizebox{\textwidth}{!}{
\begin{tabular}{@{}l l cccccc@{}} 
\toprule
& \textbf{Method}  & \textbf{3D Mask} & \textbf{2D Mask} & \textbf{Camera Interpolation} & \textbf{AP} & \textbf{AP50} & \textbf{AP25} \\ 
\midrule
\multirow{5}{*}{Replica \cite{straub2019replica}} & OpenScene \cite{peng2023openscene}   & Mask3D    & \xmark & \xmark & 10.9 & 15.6 & 17.3 \\
& OpenMask3D~\cite{takmaz2023openmask3d}            & Mask3D    & SAM  & \xmark & 13.1 & 18.4 & 24.2 \\
& Open3DIS (only 3D)       & ISBNet + SuperPoint    & \xmark & \xmark & 14.9 & 18.8 & 23.6 \\
& OpenMask3D$^\dagger$            & Mask3D    & \xmark  & \xmark & 14.9 & 18.9 & 22.5 \\
& \bf{NVSMask3D (ours)}  & Mask3D    & \xmark & \checkmark & 17.8 &21.8& 26.4\\
& \bf{NVSMask3D (ours)}      & Mask3D    & SAM & \checkmark & \textbf{18.9} & \textbf{23.9}& \textbf{27.0} \\

\midrule
\multirow{4}{*}{ScanNet++ \cite{yeshwanth2023scannet++}}
 & OpenMask3D$^\dagger$       & Mask3D    & \xmark & \xmark &  7.7 & 11.4 &  14.5 \\
  & OpenMask3D       & Mask3D    & SAM & \xmark &  11.8 & 19.2 &  25.0 \\
& \bf{NVSMask3D (ours)}     & Mask3D    & \xmark &  \checkmark& 8.3  & 12.4 & 15.7 \\
& \bf{NVSMask3D (ours)}     & Mask3D    & SAM &  \checkmark& \textbf{12.1}  & \textbf{19.6} & \textbf{25.4} \\
\bottomrule
\end{tabular}
}
\caption{Open-vocabulary 3D instance segmentation. Performance of NVSMask3D on the Replica and ScanNet++ datasets, showing a quantitative comparison with baseline methods based on OpenMask3D \cite{takmaz2023openmask3d}. Here, OpenMask3D$^\dagger$ refers to a version of OpenMask3D~\cite{takmaz2023openmask3d} that uses projected and occluded 3D masks instead of 2D masks from SAM~\cite{Kirillov_2023_ICCV}. The best results are highlighted in \textbf{bold}.}
\label{results} 
\vspace{-20pt}
\end{table*}

\subsection{Implementation details}
We use posed RGB-D pairs from the ScanNet++ and Replica datasets. For Replica, we utilize all 200 pre-processed pairs to ensure full coverage, while for ScanNet++, we uniformly sample one frame out of every five to reduce computational overhead while maintaining a representative subset. For the 3D class-agnostic proposals, we use the Mask3D~\cite{schult2023mask3d} model trained on ScanNet200 instances, relying solely on the predicted binary instance masks without additional processing through DBSCAN or other filtering algorithms. For segmented 3D instance feature assignment, we employ the CLIP~\cite{radford2021learning} visual encoder from the ViT-L/14 model, pre-trained at a 336-pixel resolution, producing features with a dimensionality of 768. For computing camera pose visibility scores, we set the depth threshold to $\delta = 0.4$ for the Replica dataset and $\delta = 0.05$ for ScanNet++, as described in~\secref{subsec:camera_visibility_score}. To select the most object-descriptive $k$ camera poses, we use $\text{top-}k = 15$ for Replica and $\text{top-}k = 5$ for ScanNet++. Unlike previous works~\cite{peng2023openscene,takmaz2023openmask3d}, we do not use text prompts such as "a {} in a scene" or other contextual phrases. Instead, we take a more direct approach by assigning each object instance its class name as the query, prioritizing visual prompts to enhance instance representation. All experiments were conducted on a workstation with an NVIDIA GeForce RTX 4090 GPU. However, for generating 3D class-agnostic proposals on ScanNet++, which requires higher GPU memory, we utilized an NVIDIA V100.


\boldparagraph{Visual prompts.} For cropping, bounding boxes are derived from projected 3D masks or, with SAM, 2D masks are generated using five sampled points from the projected 3D mask as prompts. The blurring technique follows~\cite{yang2023fine}, where valid regions are dilated and eroded to form masks. However, when the object is very small, blurring alone is insufficient to draw CLIP attention. To address this, we refined our blurring strategy by applying a 0.7 scaling factor to the larger dimension when forming square regions around the mask, ensuring the object occupies a suitable proportion of the cropped image. Similar to the baseline described in~\secref{subsubsec:baseline}, the \textit{secondary visual prompts} for view interpolation uses only the 3D mask, ensuring a fair comparison of different prompting methods applied uniformly across interpolated views, while the 2D mask is exclusively applied to non-interpolated/non-adjusted views. For non-interpolated camera poses provided in the dataset, we directly use the corresponding images instead of re-rendering them from the 3D representation, reducing synthetic image artifacts and ensuring consistency with the baselines.

\boldparagraph{Open-vocabulary 3D Instance Segmentation.} As shown in~\tabref{results}, our NVSMask3D method outperforms the baselines across all metrics on the closed-vocabulary instance segmentation task. On the Replica dataset, NVSMask3D achieves an AP of 17.8 using only 3D masks with camera interpolation, showing a significant improvement over the 3D mask-only baseline without camera interpolation (AP of 14.9). When incorporating 2D SAM masks, NVSMask3D further improves the AP to 18.9, which performs better than the 2D–3D mask baseline (AP of 13.1), demonstrating the advantage of WFB combining 2D and 3D mask information. Similarly, on the ScanNet++ dataset, our method achieves an AP of 8.3 using only 3D masks with camera interpolation, outperforming the OpenMask3D approach (AP of 7.7). With the addition of 2D SAM masks, the AP increases to 12.1, highlighting the effectiveness of our approach in leveraging multi-view and multi-modal information.
These results indicate that our camera interpolation and object-descriptive visual prompting techniques substantially enhance the CLIP feature representation for 3D objects, leading to improved performance in 3D instance segmentation tasks. The consistent improvements over both baselines confirm the effectiveness of the proposed method.

\subsection{Ablation Study \& Analysis}
\begin{table}[t]
    \centering
    \begin{minipage}{0.38\textwidth} 
        \captionof{table}{\textbf{Ablations of secondary visual prompts}. Comparison of different visual prompts with our external visual prompting (camera interpolation) method without SAM and WFB $(\alpha = 0.5)$.} 
        \label{visual_prompt}
    \end{minipage}
    \begin{minipage}{0.6\textwidth} 
        \centering
        \begin{tabular}{@{}lcccccc@{}}
            \toprule
            Visual prompt & Interp & AP & AP50 & AP25 \\ 
            \midrule
            Cropping & \xmark & 14.9 & 18.9 & 22.5 \\
            Cropping & \checkmark & 16.1 & 20.0 & 24.3 \\
            Blur Reverse Mask & \checkmark & 15.6 & 19.4 & 23.5 \\
            Segmented Gaussian (Ours) & \checkmark & \textbf{17.8} & \textbf{21.8} & \textbf{26.4} \\
            \bottomrule
        \end{tabular}
    \end{minipage}%
    \vspace{-10pt}
\end{table}
\begin{wraptable}{l}{0.6\textwidth}  
    \centering
    \vspace{-20pt}
    \begin{tabular}{@{}lccccc@{}}
        \toprule
        Feature Fusion & $N$ Interpolation & AP & AP50 & AP25 \\ 
        \midrule
        average & \xmark & 14.9 & 18.9 & 22.5 \\
        average & 1 & 16.1 & 20.0 & 24.3 \\
        average & 2 & 14.5 & 18.0 & 22.1 \\
        average & 3 & 13.6 & 16.8 & 20.4 \\
        \midrule
        WFB$(\alpha = 1)$ & 1 & 16.1 & 20.0 & 24.3 \\
        WFB$(\alpha = 1)$ & 2 & 15.9 & 19.8 & 24.0 \\
        WFB$(\alpha = 1)$ & 3 & 15.9 & 19.8 & 24.0 \\
        WFB$(\alpha = 0.5)$ & 1 & 17.8 & 21.8 & 26.4 \\
        WFB$(\alpha = 0.5)$ & 2 & 17.6 & 21.6 & 26.2 \\
        WFB$(\alpha = 0.5)$ & 3 & 17.8 & 21.8 & 26.4 \\
        \bottomrule
    \end{tabular}
    \caption{\textbf{Amblations of different feature futsion}. Comparison of the proposed averaging and weighted feature balancing (WFB) with $N$ cameras interpolated between 2 cameras with cropping visual prompt without SAM.} 
    \label{WFB}
    \vspace{-70pt}
\end{wraptable}

\subsubsection{Impact of secondary visual prompts } To validate the design choices of our method, a series of ablation studies are conducted on the Replica dataset.
The proposed approach builds on top-$k$ selected images with cropping by introducing interpolated camera views and applying various visual prompting techniques, including segmented Gaussian, cropping, and blur reverse mask.  This addition of interpolated views enhances the descriptiveness of the prompts, resulting in stronger object representations and improved segmentation, as presented in~\tabref{visual_prompt}. Specifically, we evaluate a simple cropping-based prompt without camera interpolation, which achieves an AP of 14.9. Adding camera interpolation to cropping raises the AP to 16.1, highlighting the benefit of additional object-descriptive views. Similarly, the blur reverse mask approach with camera interpolation outperforms basic cropping, achieving an AP of 15.6. The best performance is achieved by combining camera interpolation with segmented Gaussian prompts, yielding an AP of 17.8. This result indicates that segmented Gaussian prompting, coupled with interpolated views, effectively emphasizes the object while reducing background distractions, enhancing the CLIP feature representation of the original top-$k$ cropped images. These results demonstrate that using additional object-descriptive visual prompts improves segmentation accuracy. Furthermore, combining the neighboring context (\cf \textit{Cropping} in~\tabref{visual_prompt}) CLIP feature with isolated object-focused (\cf \textit{Segmented Gaussian} in~\tabref{visual_prompt}) CLIP feature can further improve performance.

\boldparagraph{Impact of Different Feature Fusion. }To assess the impact of different feature fusion methods and the number of interpolated views, we conducted additional ablation studies, with results summarized in~\tabref{WFB}. We compared simple averaging with our proposed weighted feature balancing (WFB) approach, testing various numbers of $N_{\text{interp}}$ between two selected camera poses. The results demonstrate that directly averaging features from NVS-generated images introduce noise, especially as the number of interpolated views increases. Simple averaging fails to properly balance contributions from interpolated views, resulting in a performance drop as $N_{\text{interp}}$ grows. In contrast, our WFB method effectively addresses this issue by adjusting the feature contribution from each interpolated view, preventing the performance drop seen with basic averaging as $N_{\text{interp}}$ increases. Additionally, we experimented with different values of the contribution parameter $\alpha$ in WFB. Across all tested $\alpha$ values, WFB successfully mitigates the underperformance caused by an increasing number of interpolated views. However, we observed that a moderate $\alpha$ value, such as $\alpha = 0.5$, produced the best performance, achieving a higher AP. This indicates that carefully selecting $\alpha$ can further enhance the model’s ability to balance contributions from multiple views effectively.


\section{Discussion}
Our results highlight the effectiveness of NVSMask3D, particularly through the integration of camera interpolation and segmented Gaussian as object descriptive visual prompts. These techniques effectively emphasize object features while minimizing irrelevant background noise, leading to improved CLIP feature representations for 3D objects and enhanced segmentation accuracy. The addition of WFB further stabilizes performance by adjusting the influence of interpolated views, addressing potential noise that can arise from direct feature averaging. Notably, our findings in object descriptive visual prompting indicate that hard visual prompts, traditionally used in 2D, can also be highly effective within 3D applications, without requiring any joint optimization of 2D and 3D features or finetuning.

Compared to baseline methods, NVSMask3D consistently achieves higher AP metrics on challenging datasets like Replica and ScanNet++. By leveraging camera interpolation as an external visual prompt specifically for 3D, our method not only enhances segmentation performance but also provides a flexible framework capable of generalizing across diverse 3D scenes without retraining. This adaptability underscores NVSMask3D’s potential for other real-world open vocabulary applications, where data varies significantly across scenes and objects.

\textbf{Acknowlegement} We thank Matias Turkulainen and Hanlin Yu for their valuable contributions, including coding and insightful discussions. We also thank Songyou Peng for his fruitful input. We also acknowledge funding from the Research Council of Finland (grants 352788, 353138, 362407) and CSC and Aalto-IT for the computational resources.

{
 \bibliographystyle{splncs04}
 \bibliography{bibliography}
 }
%




\clearpage
\appendix

\section{Computational Efficiency Breakdown}
We evaluate the computational efficiency of NVSMask3D by measuring the execution time of its core components on the Replica Office0 scene. The 3D mask proposal step takes 13 seconds. Optimizing 3D-GS for 10,000 steps, using the default settings from Nerfstudio~\cite{tancik2023nerfstudio,ye2024gsplat}, adds approximately 3 minutes. Geometric median estimation and view interpolation introduce a minor overhead of up to 1 second per instance. CLIP inference step remains highly efficient, requiring only 1–2 ms per query. The main computational overhead, compared to our baselines, comes from 3D-GS optimization and camera interpolation, including per-instance geometric median calculation and view rendering. Despite these additional steps, the overall framework remains computationally feasible for real-world applications.

\section{Details of Geometric Median}
\label{sec:geometric_median}
As defined in Equation~\eqref{eq:geometric_median}, the geometric median minimizes the sum of distances to all points. We compute the geometric median of a 3D object using an optimized version of Weiszfeld's algorithm, which ensures numerical stability and efficient convergence. 
The algorithm terminates after a maximum of 1000 iterations or when the improvement falls below a convergence threshold \( 10^{-5} \). 


\begin{algorithm}[h!]
\caption{Optimized Weiszfeld Algorithm for Geometric Median}
\label{alg:weiszfeld}
\KwIn{Set of 3D points $\mathbf{P} = \{\mathbf{p}_i \in \mathbb{R}^3 \}_{i=1}^N$, 
       convergence ratio \texttt{eps\_ratio = \(10^{-5}\)}, 
       maximum iterations \texttt{max\_iter = 1000}}
\KwOut{Geometric Median $\mathbf{c} \in \mathbb{R}^3$}

\textbf{Initialize:} \\
Compute data scale $\sigma = \text{mean}(\text{std}(\mathbf{P}, \text{axis}=0))$; \\
Set convergence threshold $\epsilon = \sigma \cdot \texttt{eps\_ratio}$; \\
Set initial estimate $\mathbf{c} = \frac{1}{N} \sum_{i=1}^N \mathbf{p}_i$; \\

\For{$t = 1$ \textbf{to} \texttt{max\_iter}}{
    Compute distances $d_i = \|\mathbf{p}_i - \mathbf{c}\|_2$ for all $i$; \\
    Mask valid points to avoid division by zero: $\mathbf{M} = \{i \mid d_i > 10^{-10}\}$; \\
    \If{$\mathbf{M}$ is empty}{
        \textbf{break} \tcp*[l]{All points converge
to $\mathbf{c}$}
    }
    Compute weights $w_i = \frac{1}{d_i}$ for $i \in \mathbf{M}$; \\
    Update $\mathbf{c}_{\text{new}} = \frac{\sum_{i \in \mathbf{M}} w_i \cdot \mathbf{p}_i}{\sum_{i \in \mathbf{M}} w_i}$; \\
    \If{$\|\mathbf{c}_{\text{new}} - \mathbf{c}\|_2 < \epsilon$}{
        \textbf{break} \tcp*[l]{Convergence achieved}
    }
    $\mathbf{c} \gets \mathbf{c}_{\text{new}}$;
}
\Return $\mathbf{c}$;
\end{algorithm}
\clearpage
\section{Qualitative result}
\begin{figure}[h!]
\vspace{-1em}
    \centering
    \begin{subfigure}{\textwidth}
        \centering
        \begin{minipage}{0.329\textwidth}
            \centering
            \includegraphics[width=\textwidth]{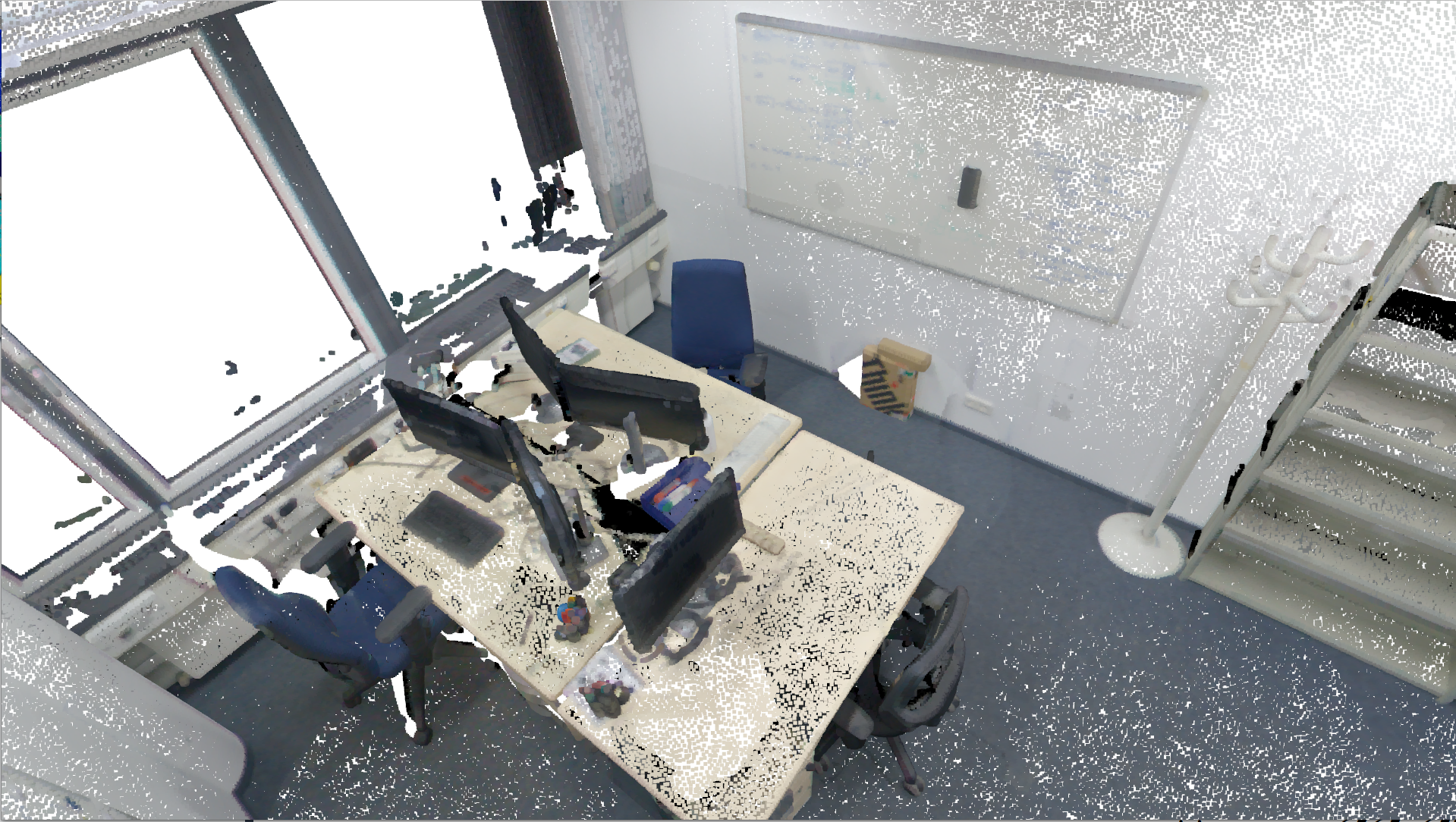} 
            \caption{Affordance}
        \end{minipage}
        \hfill
        \begin{minipage}{0.329\textwidth}
            \includegraphics[width=\textwidth]{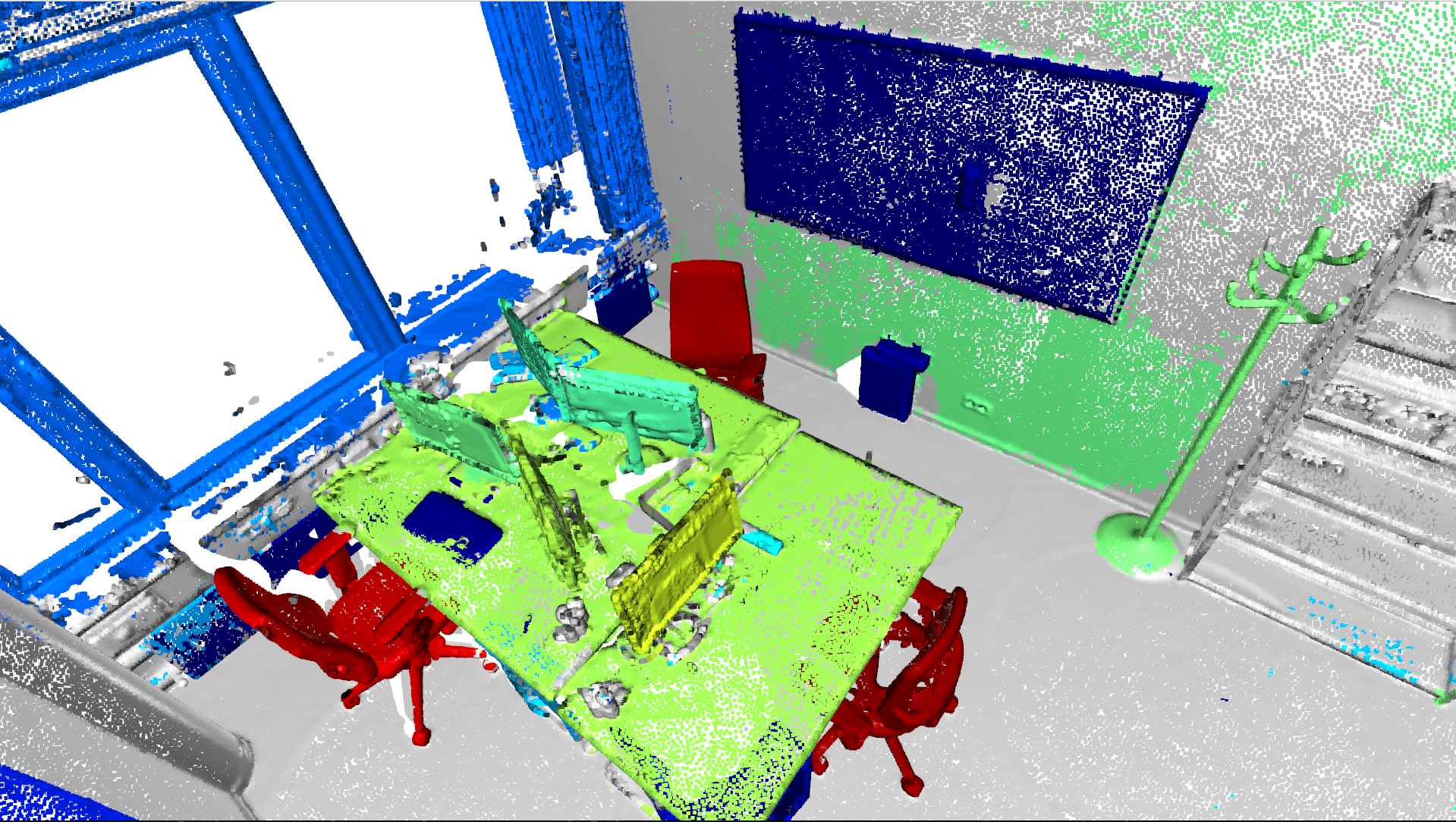} 
            \caption{\textit{"sit"}}
        \end{minipage}
        \hfill
        \begin{minipage}{0.329\textwidth}
            \includegraphics[width=\textwidth]{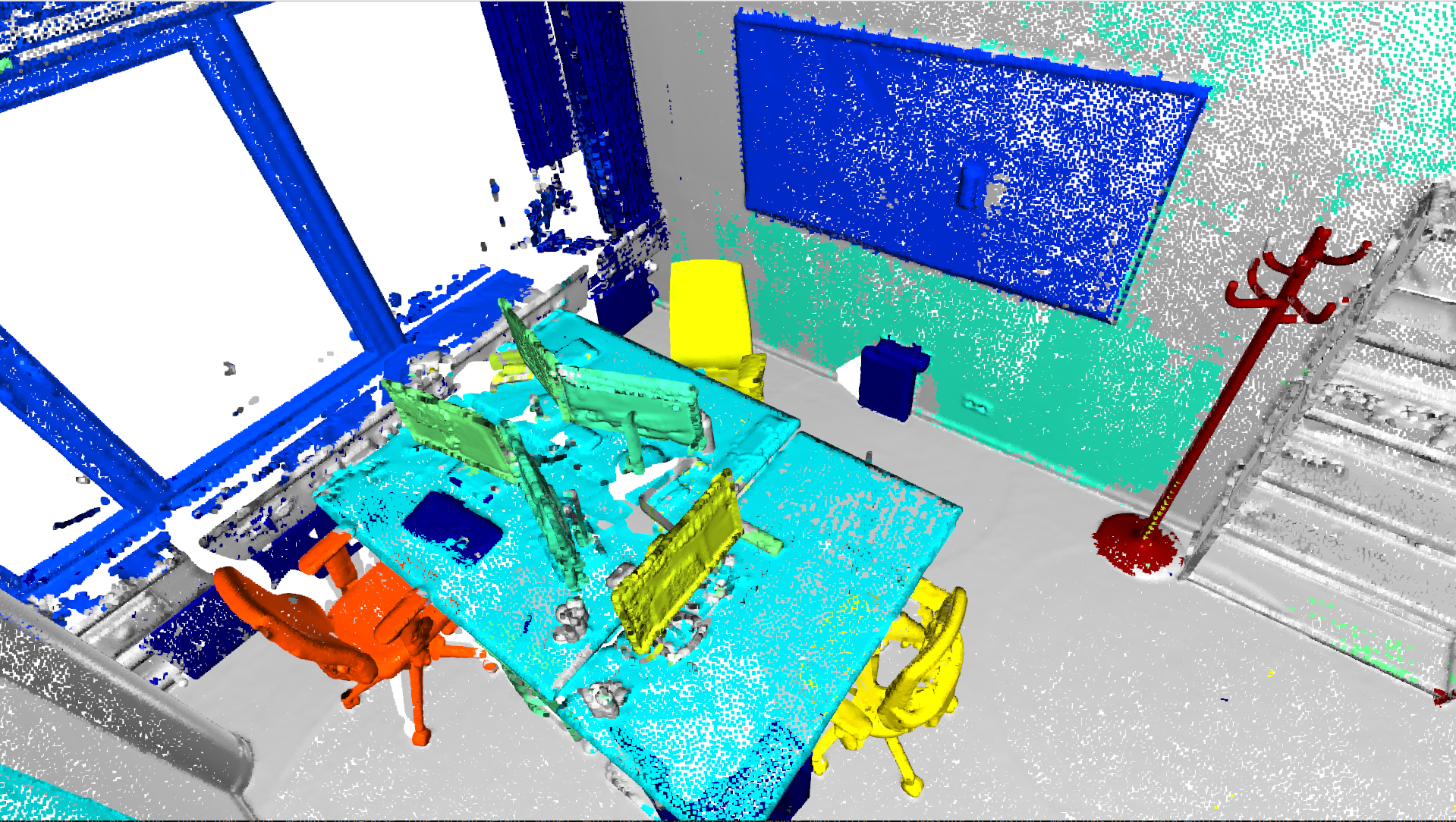} 
            \caption{\textit{"hang"}}
        \end{minipage}
    \end{subfigure}

    \begin{subfigure}{\textwidth}
        \centering
        \begin{minipage}{0.329\textwidth}
            \centering
            \includegraphics[width=\textwidth]{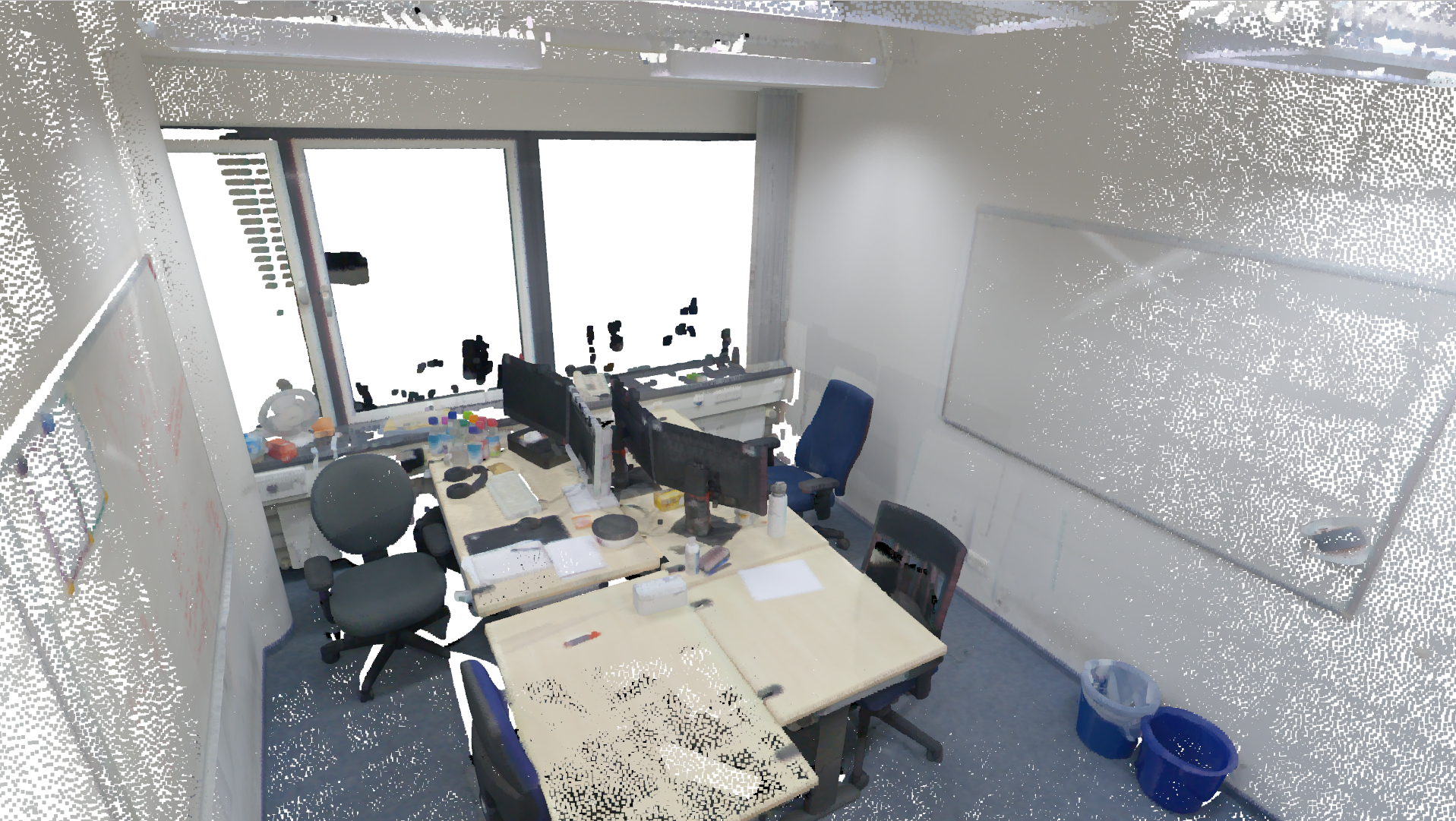} 
            \caption{States}
        \end{minipage}
        \hfill
        \begin{minipage}{0.329\textwidth}
            \includegraphics[width=\textwidth]{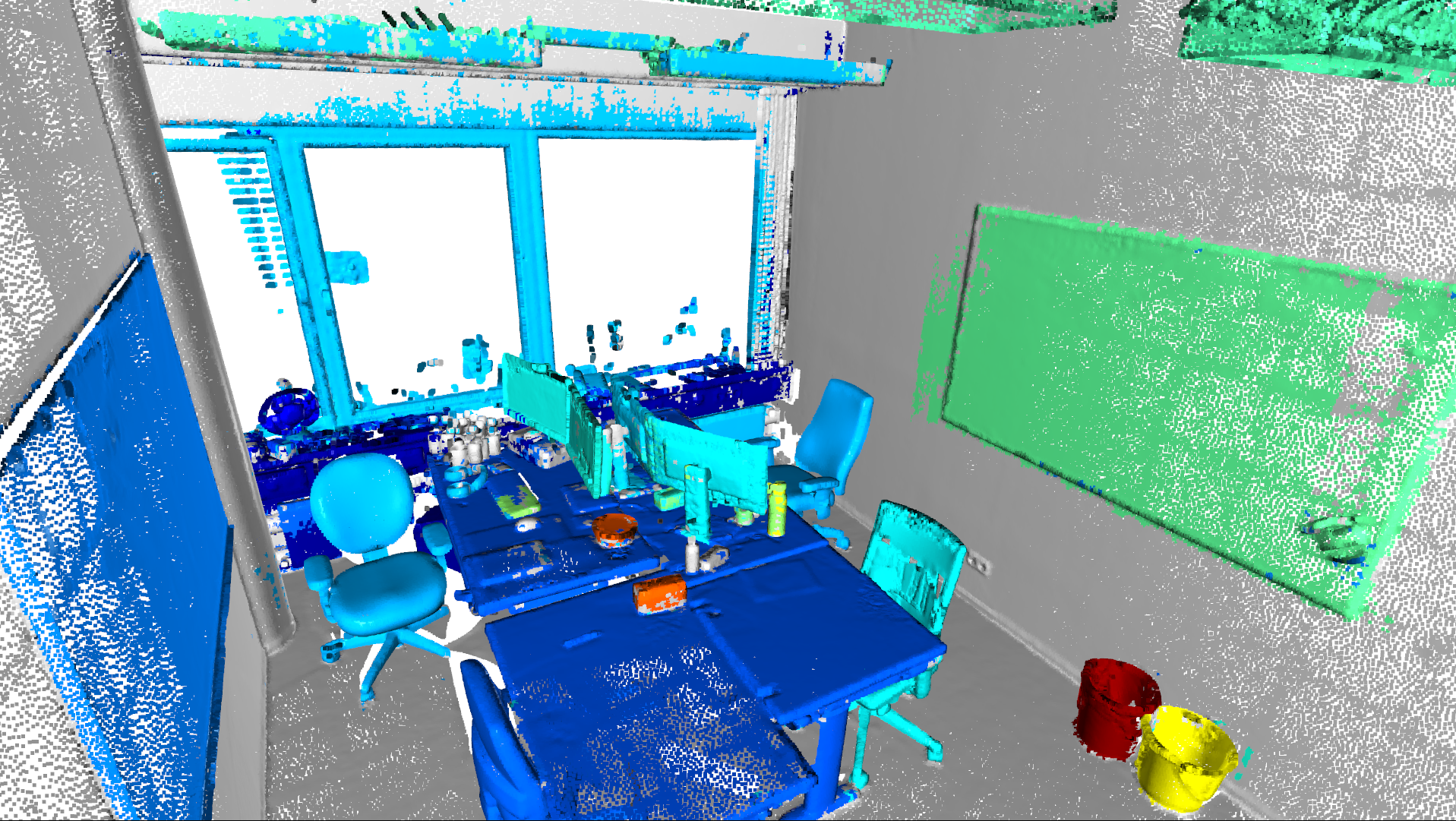} 
            \caption{\textit{"a bin with a white bag"}}
        \end{minipage}
        \hfill
        \begin{minipage}{0.329\textwidth}
            \includegraphics[width=\textwidth]{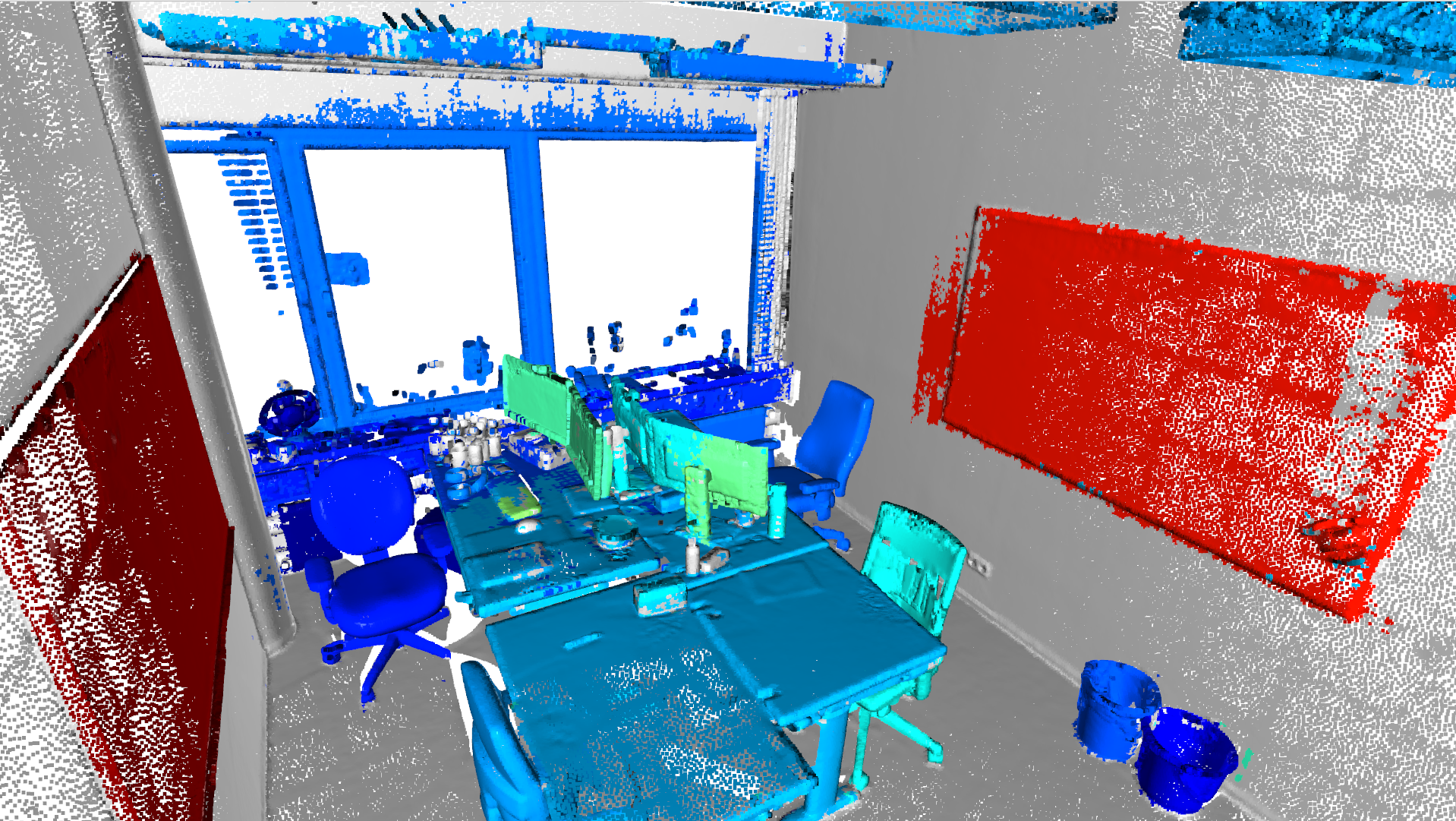} 
            \caption{\textit{"a written whiteboard"}}
        \end{minipage}
    \end{subfigure}

        \begin{subfigure}{\textwidth}
        \centering
        \begin{minipage}{0.329\textwidth}
            \centering
            \includegraphics[width=\textwidth]{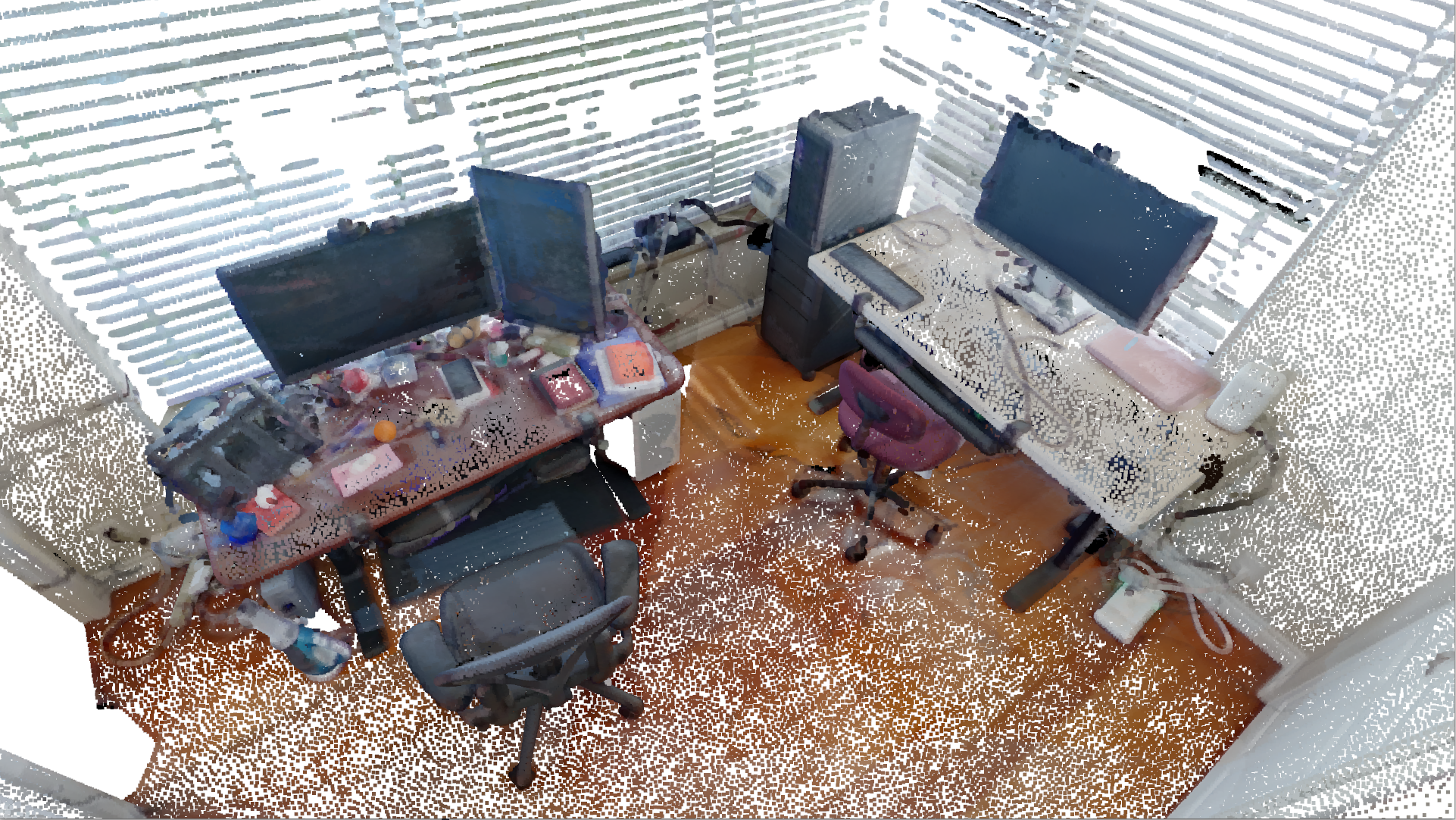} 
            \caption{Color}
        \end{minipage}
        \hfill
        \begin{minipage}{0.329\textwidth}
            \includegraphics[width=\textwidth]
            {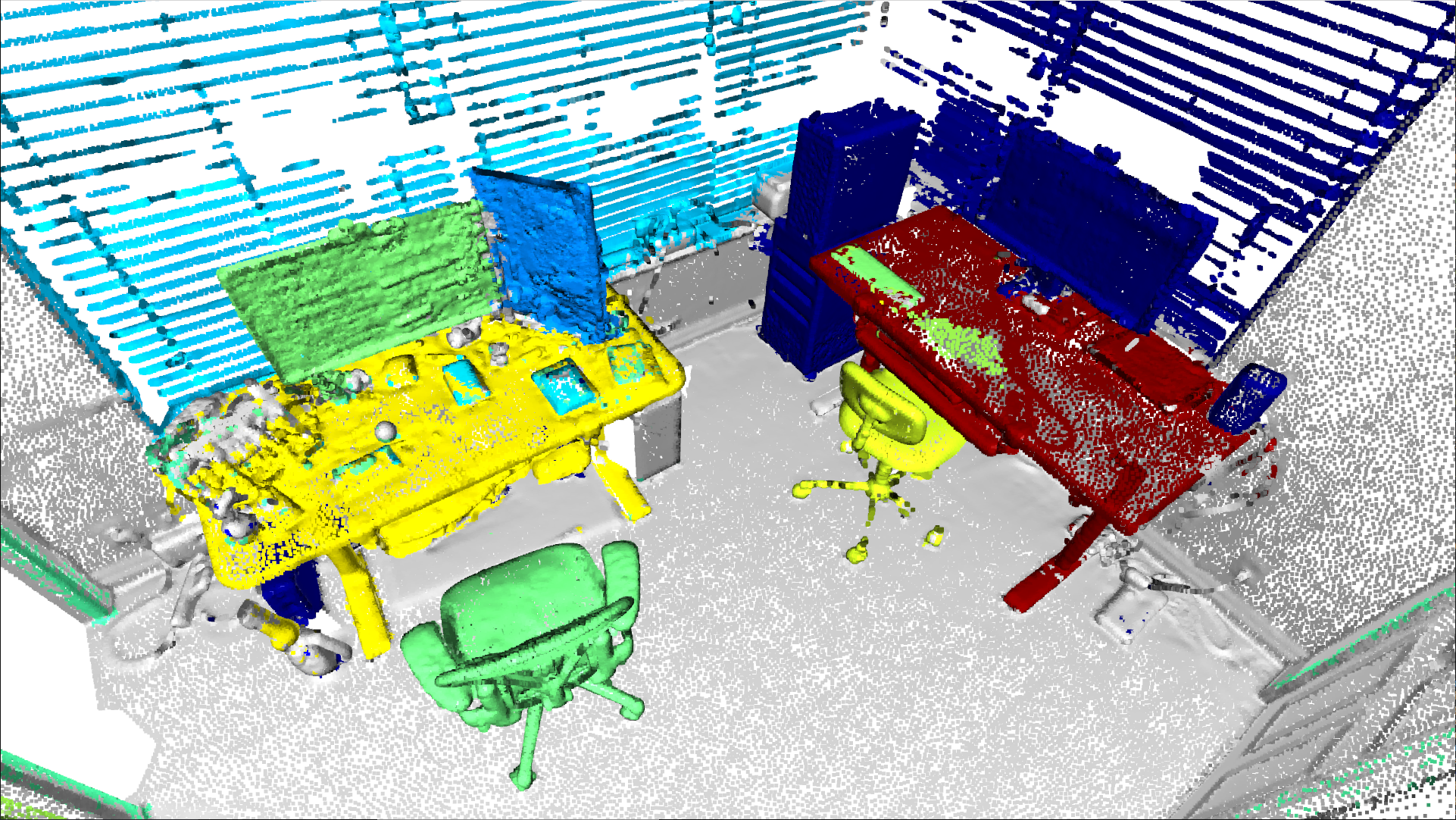} 
            \caption{\textit{"a white desk"}}
        \end{minipage}
        \hfill
        \begin{minipage}{0.329\textwidth}
            \includegraphics[width=\textwidth]
            {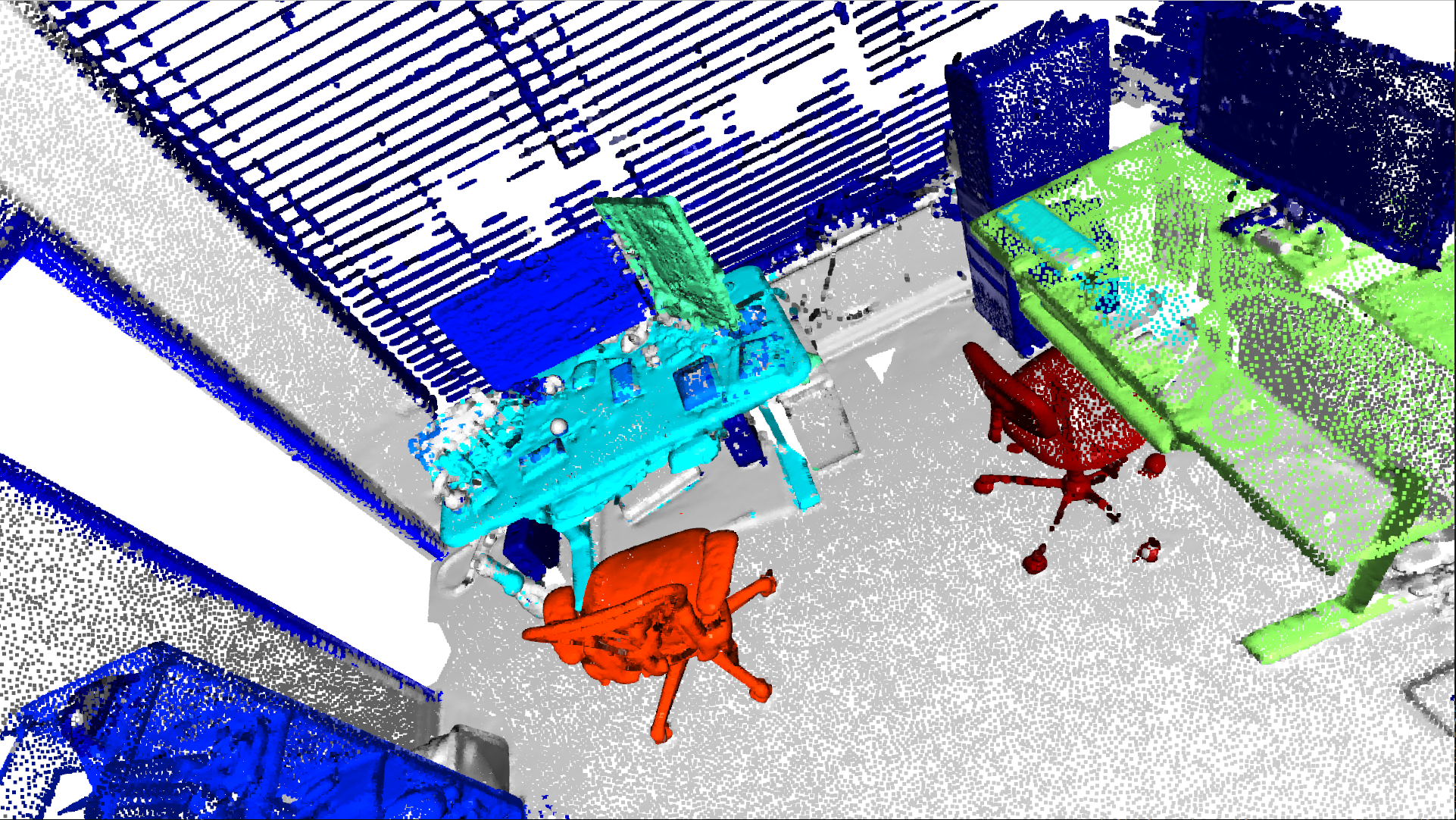} 
            \caption{\textit{"a red chair"}}
        \end{minipage}
    \end{subfigure}

            \begin{subfigure}{\textwidth}
        \centering
        \begin{minipage}{0.329\textwidth}
            \centering
            \includegraphics[width=\textwidth]{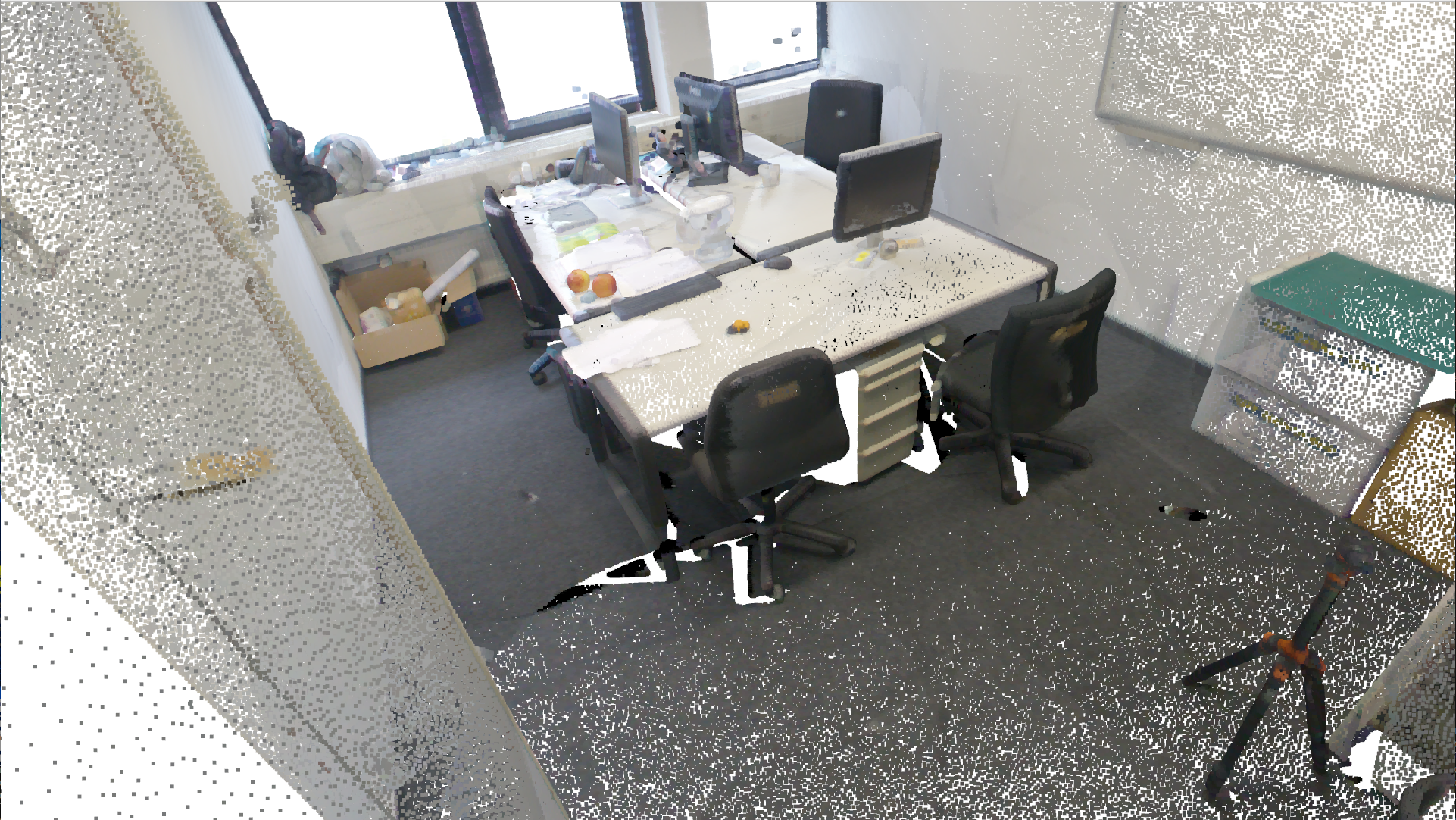} 
            \caption{Activity}
        \end{minipage}
        \hfill
        \begin{minipage}{0.329\textwidth}
            \includegraphics[width=\textwidth]{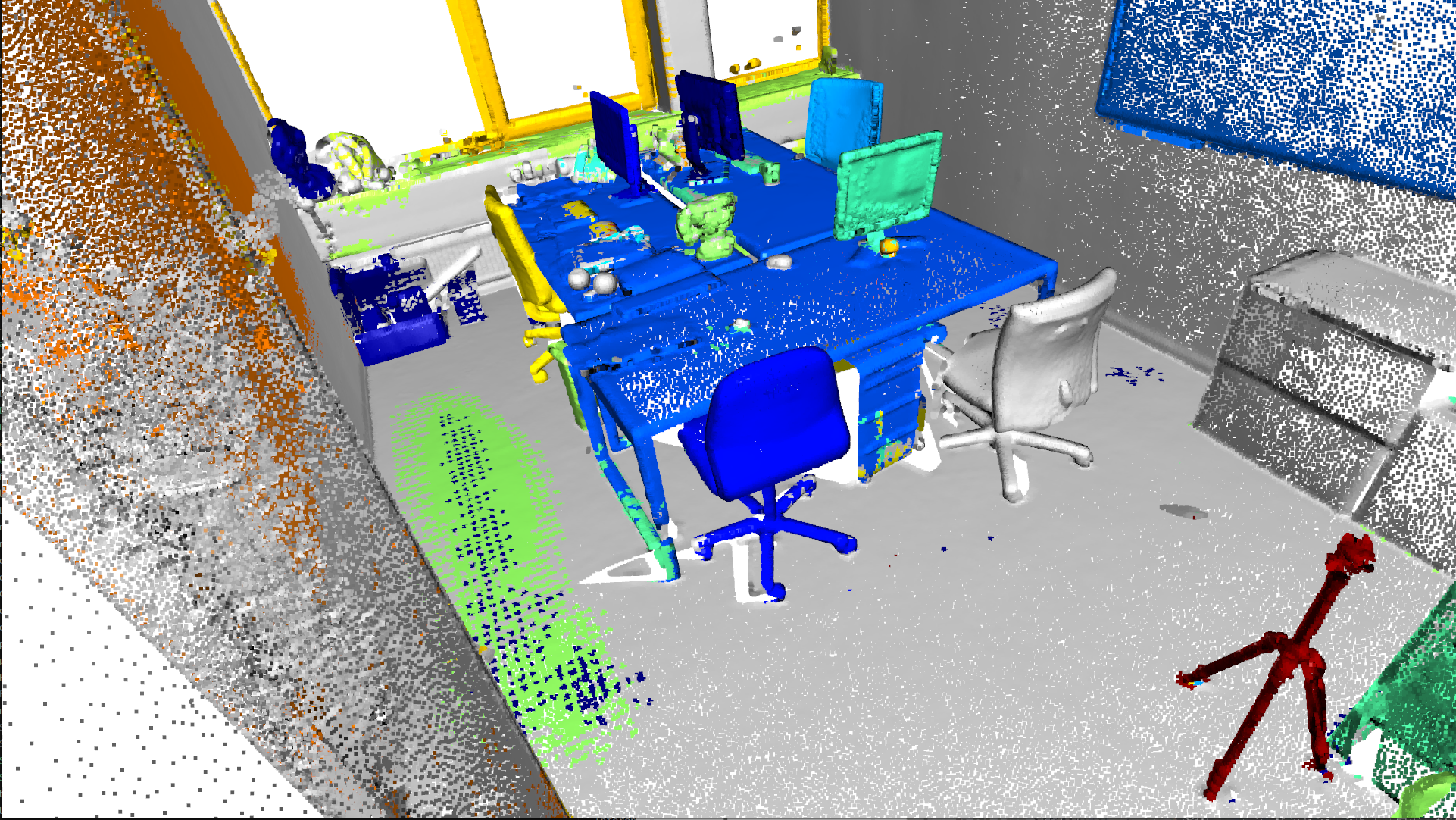} 
            \caption{\textit{"photographing"}}
        \end{minipage}
        \hfill
        \begin{minipage}{0.329\textwidth}
            \includegraphics[width=\textwidth]{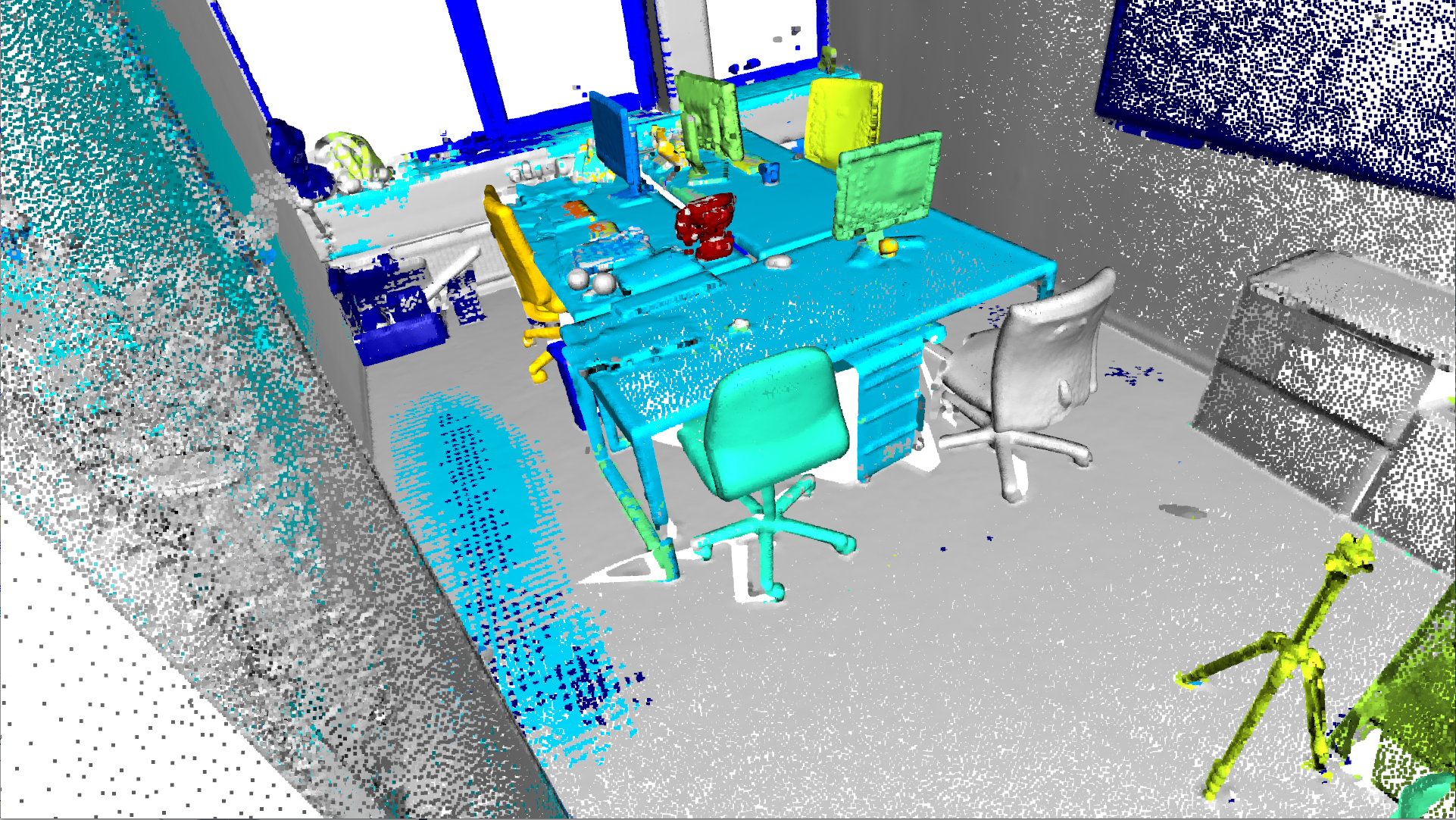} 
            \caption{\textit{"drinking"}}
        \end{minipage}
    \end{subfigure}

    \begin{subfigure}{\textwidth}
        \centering
        \begin{minipage}{0.329\textwidth}
            \centering
            \includegraphics[width=\textwidth]{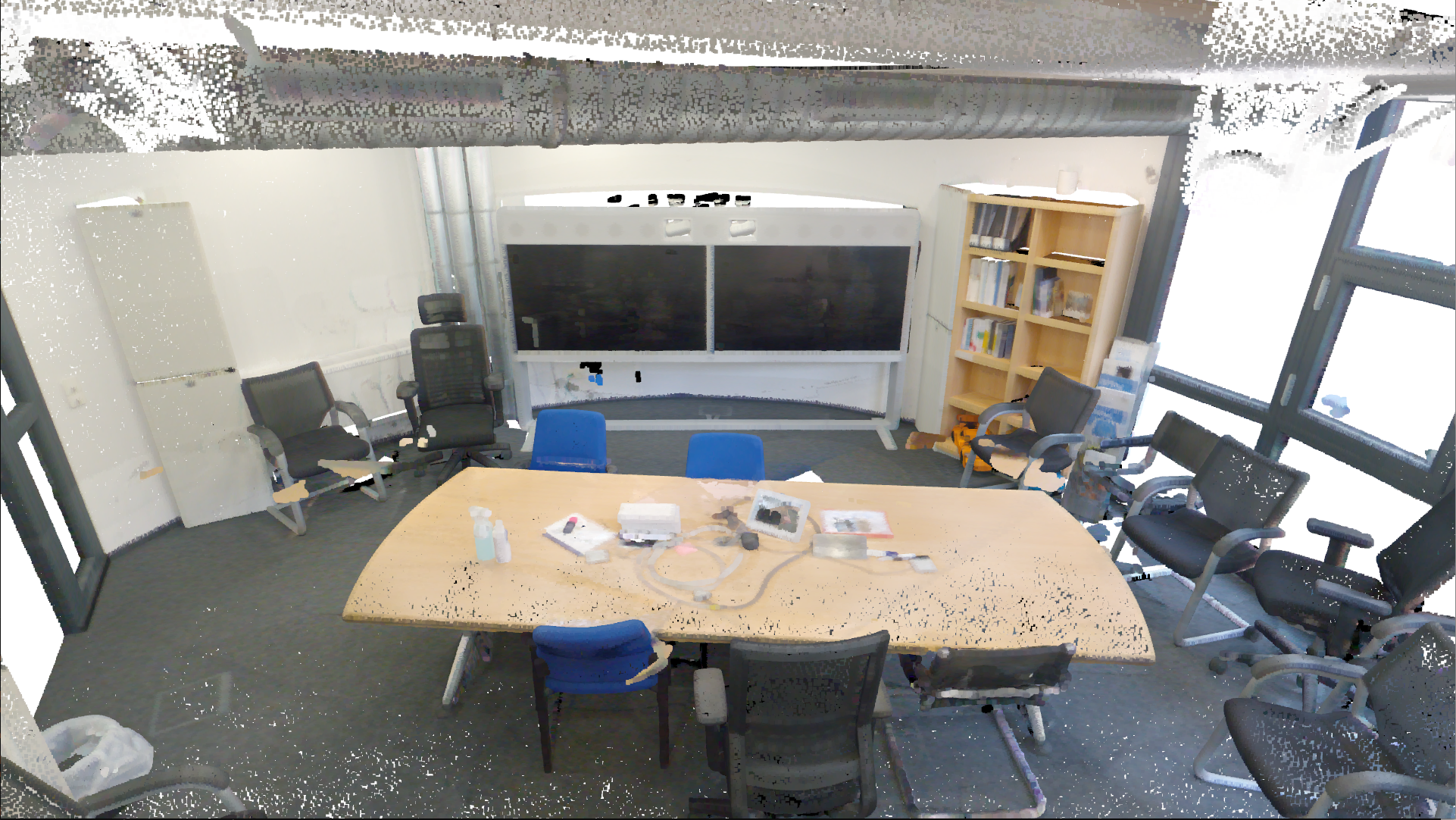} 
            \caption{Object}
        \end{minipage}
        \hfill
        \begin{minipage}{0.329\textwidth}
            \includegraphics[width=\textwidth]{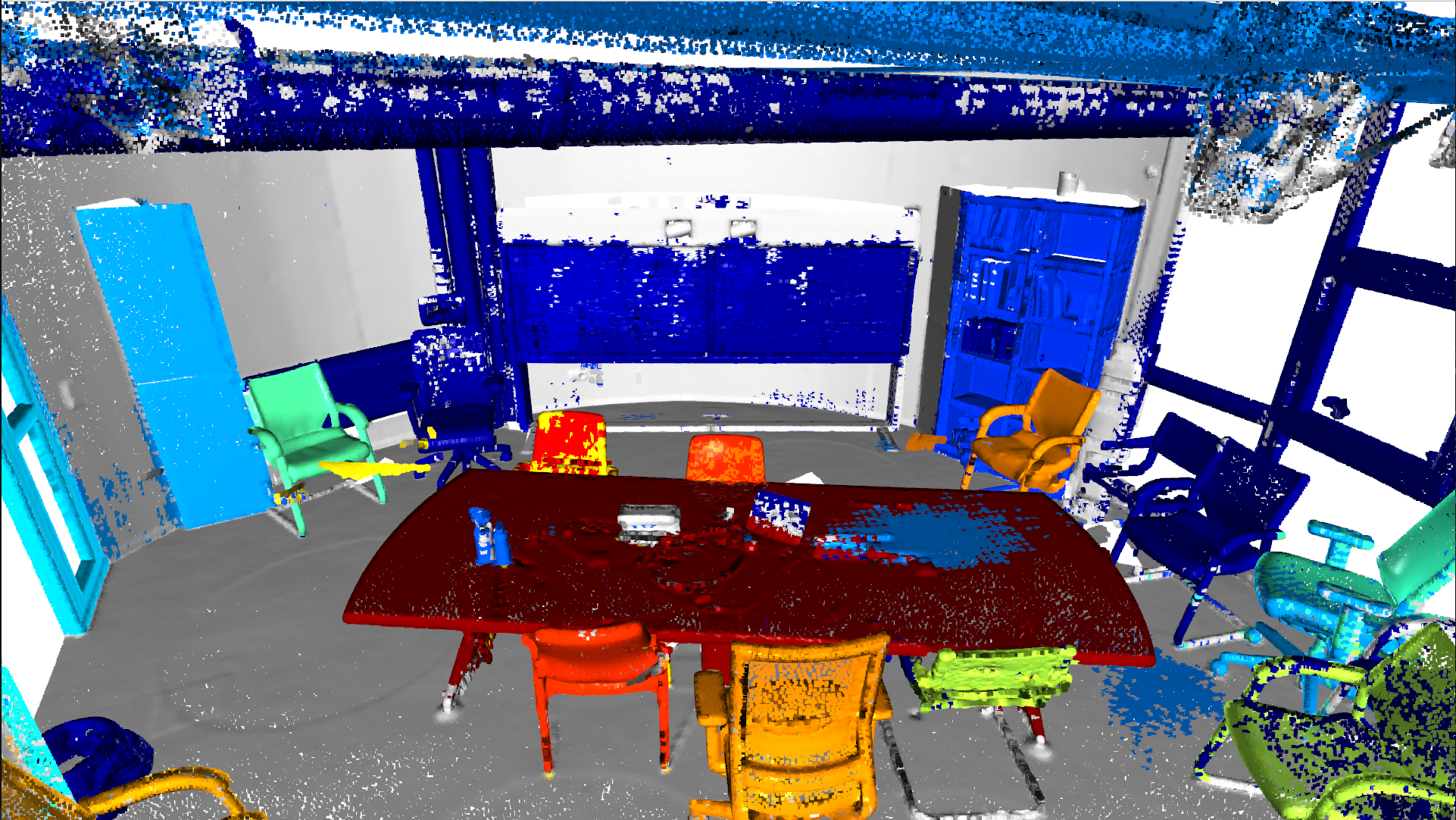} 
            \caption{\textit{"rectangular table"}}
        \end{minipage}
        \hfill
        \begin{minipage}{0.329\textwidth}
            \includegraphics[width=\textwidth]{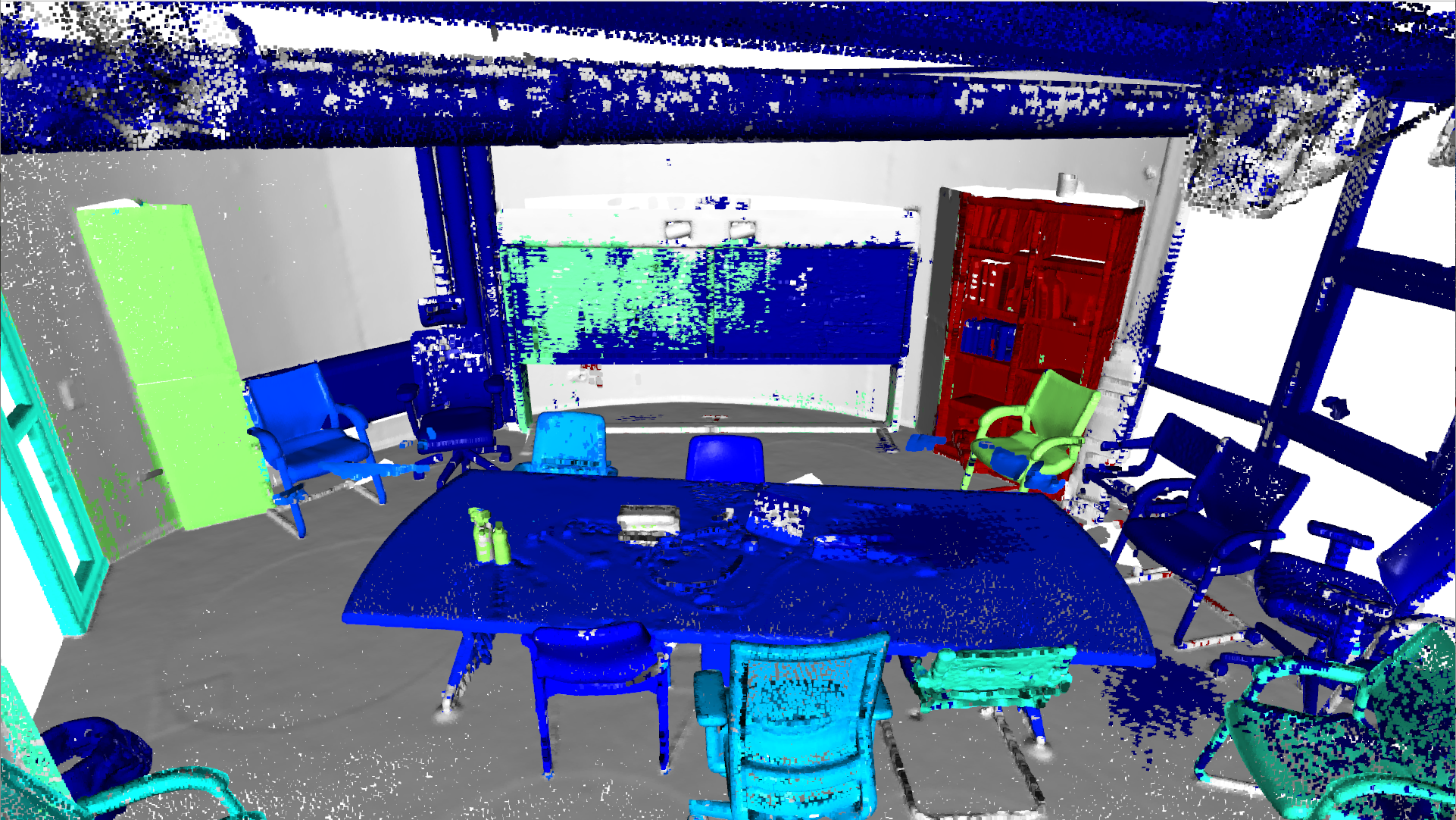} 
            \caption{\textit{"shelf with items"}}
        \end{minipage}
    \end{subfigure}
    
    \caption{\textbf{Semantic Segmentation and Retrieval in Indoor Scenes.} This figure showcases NVSMask3D's open-vocabulary instance segmentation results across various query categories: \textbf{(a)} Affordance (\textit{"sit"}, \textit{"hang"}), \textbf{(d)} States (\textit{"a bin with a white bag"}, \textit{"a written whiteboard"}), \textbf{(g)} Colors (\textit{"a white desk"}, \textit{"a red chair"}), \textbf{(j)} Activities (\textit{"photographing"}, \textit{"drinking"}), and \textbf{(m)} Objects (\textit{"rectangular table"}, \textit{"shelf with items"}). Each row includes the original point cloud (left) and segmentation results with highlighted queries (middle, right).}
    \label{fig:qualitative_our}
\end{figure}

\end{document}